\pdfoutput=1
\documentclass[5p,times]{elsarticle} 

\usepackage{times}
\usepackage{epsfig}
\usepackage{graphicx}
\usepackage{amsmath}
\usepackage{amssymb}
\usepackage{algorithm,algpseudocode,amsmath}
\usepackage{bm,upgreek}

\usepackage{url}
\usepackage{subcaption}
\usepackage{bm}
\usepackage{multirow}
\usepackage{longtable}

\newcommand{\argmax}{\mathrm{argmax}}

\makeatletter

\let\oldReturn\Return
\renewcommand{\Return}{\State\oldReturn}

\makeatletter
\newcounter{phase}[algorithm]
\newlength{\phaserulewidth}
\newcommand{\setphaserulewidth}{\setlength{\phaserulewidth}}
 \makeatother

\journal{Pattern Recognition}

\setphaserulewidth{.7pt}
\begin{document}

\begin{frontmatter}

\title{Preferences Prediction using a Gallery of Mobile Device based on Scene Recognition and Object Detection}
\author[1]{A.V. Savchenko}
\ead{avsavchenko@hse.ru}

\author[2]{K.V. Demochkin}
\ead{kdemochkin@gmail.com}

\author[1,2]{I.S. Grechikhin}
\ead{gis1093@mail.ru}

\address[1]{HSE University\\
Laboratory of Algorithms and Technologies for Network Analysis, Nizhny Novgorod, Russia
}
\address[2]{St. Petersburg Department of Steklov Institute of Mathematics\\
Samsung-PDMI Joint AI Center, St. Petersburg, Russia}

\begin{abstract}
In this paper user modeling task is examined by processing a gallery of photos and videos on a mobile device. We propose novel engine for user preference prediction based on scene recognition, object detection and facial analysis. At first, all faces in a gallery are clustered and all private photos and videos with faces from large clusters are processed on the embedded system in offline mode. Other photos may be sent to the remote server to be analyzed by very deep models. The visual features of each photo are obtained from scene recognition and object detection models. These features are aggregated into a single user descriptor in the neural attention block. The proposed pipeline is implemented for the Android mobile platform. Experimental results with a subset of Photo Event Collection, Web Image Dataset for Event Recognition and Amazon Fashion datasets demonstrate the possibility to process images very efficiently without significant accuracy degradation. The source code of Android mobile application is publicly available at \url{https://github.com/HSE-asavchenko/mobile-visual-preferences}.
\end{abstract}


\end{frontmatter}



\section{Introduction}
\label{sec:1}
Important features of today's mobile devices are personalized services that adapt to individual users and collect and model user-specific information. Personalized recommendations are developed to assist customers in finding relevant things within large item collections. The design of such systems requires the careful consideration of user modeling approach, which defines a user's interest in his or her profile. Conventional content-based recommender systems use only structured and textual data~\cite{yu2019cross}. However, a large gallery of photos is usually available on a mobile device, which can be also used for understanding of such interests as sport, gadgets, fitness, cloth, cars, food, pets, etc. The usage of contemporary pattern recognition methods for photos from the gallery has many advantages for improving the quality of recommender system.

It is necessary to emphasize that the processing of photos significantly differs from above-mentioned applications due to the need for protections of users' privacy~\cite{yang2020graph}. The gallery on a mobile phone usually contains much more photos when compared to the number of photos in their publicly available profile in social networks. Hence, the analytics engines should be preferably run on the embedded system, because most photos contain private information, for which the user may not permit processing on the remote server. As a result, the state-of-the-art very deep convolutional neural networks (CNNs)~\cite{goodfellow2016deep} cannot be directly applied due to their enormous inference time and energy consumption. Thus, in this paper we consider several directions for faster prediction of the user preferences, namely, scene recognition~\cite{farinella2015representing,zhou2018places} in order to extract such interests as art and theaters, nightlife, sport, etc.; detection of objects~\cite{huang2017speed,grechikhin2019user} including food, pets, musical instruments, vehicles, brand logos~\cite{su2020scalable}; analysis of demography and sociality by processing facial images in photos and videos including facial clustering~\cite{savchenko2019efficient}.

This paper makes two main contributions to efficient recognition of photos and videos. Firstly, we propose a technological framework to predict user preferences given photos on a mobile device that exploits as much as possible offline processing on the personal device. Secondly, we describe a representation of image by using scores and embeddings from the scene recognition model combined with the scores of object detector. We demonstrate that such a representation is suitable for several different pattern recognition tasks including visual product recommendation and event recognition~\cite{wang2018transferring}. In particular, we propose to combine obtained representations of all photos from a gallery into a single descriptor of a user by modifying the neural aggregation module with an attention mechanism originally used in video-based face recognition~\cite{yang2017neural} in order to decrease its running time without degradation in accuracy.

The rest of the paper is organized as follows. In Section~\ref{sec:2}, we review work related to our task. In Section~\ref{sec:3}, we present the proposed pipeline for inferring user's profile by processing a set of photos. In Section~\ref{sec:4}, the trade-off between accuracy and complexity of various CNN-based models are experimentally studied. Finally, concluding comments are given in Section~\ref{sec:5}.

\section{Literature survey}
\label{sec:2}

\subsection{Image, scene and event recognition}
The majority of computer vision literature focus on the problem of object recognition or scene recognition, partially due to the simplicity of object and scene concepts~\cite{wang2018transferring} and the availability of large-scale datasets, e.g., ImageNet and Places. Such models are known to be used in recognition of more complex images with large variations in visual appearance and structure, e.g., events that captures the complex behavior of a group of people, interacting with multiple objects, and taking place in a specific environment (holidays, sport events, wedding, activity, etc.)~\cite{wang2018transferring}. 

Nowadays deep CNNs provide an excellent accuracy in many large-scale image classification problems~\cite{goodfellow2016deep,szegedy2017inception}. Various recent studies have shown that the increase in the number of layers (depth) lead to much more accurate solutions. The ImageNet challenge highlights this trend to deeper models. For example, the winner of 2012 edition (AlexNet) contains only 8 layers, though the 2015 winner ResNet consists of 152 layers, while InceptionResNet v2 model has more than 450 layers~\cite{szegedy2017inception}. The increase of the number of layers leads to increase of the the running time of inference in contemporary deep models. As a result, this time may be too high for real-time processing especially if expensive GPU is not available. The need for improving the speed of CNNs has become evident several years ago when it was shown that the deeper is the model, the more accurate are its results. The most remarkable research direction in this field is the optimization of algorithms and neural network architectures The CNN compression~\cite{han2015deep} includes the usage of pruning in which connections between neurons with low weights are removed and the network is fine-tuned, until achieving the allowable decrease in accuracy. Such pruning reduces the model size but does not lead to significant inference speed-up. However, there exist various structural pruning methods~\cite{mittal2018recovering,molchanov2016pruning}, which removes entire convolutional channels. 
There exist also matrix decomposition methods, which decompose the computation of weight matrices (or tensors) in each layers as a sequence of operations with lower order matrices and vectors~\cite{grachev2019}. These methods allow to explicitly choose between the space complexity and the accuracy gain by setting the decomposition rank.

In order to recognize complex images, several CNNs may be combined. For example, the paper~\cite{wang2018transferring} introduces an ensemble of two CNNs trained to classify objects from ImageNet and scenes from Places dataset. Four different layers of fine-tuned CNN were used to extract features and perform Linear Discriminant Analysis to obtain the top entry in the ChaLearn LAP 2015 cultural event recognition challenge~\cite{rothe2015dldr}.

\subsection{Object detection}
A special interest for user modeling is involved in the CNN-based object detection, which can discover particular categories of interests, e.g., interior objects, food, transport, sports equipment, animals, etc. This model are much more computationally difficult than the above-mentioned CNNs used for image recognition, because they contain additional structures to transfer visual representation obtained by the CNN into predictions of multiple object positions and confidence scores. That is why there is a significant demand for developing efficient architectures of CNNs~\cite{huang2017speed}, which can be implemented directly on a mobile device. There exist a number of architectures that computationally effective and at the same time have good accuracy: SSD (Single shot detector)~\cite{ssd}, SSDLite and YOLO (You Only Look Once) with different variations of MobileNet~\cite{sandler_inverted_2018} in a backbone CNN. Unfortunately, if it is necessary to detect small objects (road signs, food, fashion accessories, etc.), the accuracy of such computationally efficient detectors is usually much lower when compared to Faster R-CNN~\cite{ren2015faster} with very deep backbone CNN, such as ResNet or InceptionResNet~\cite{szegedy2017inception}. It is important to emphasize that there exists several papers devoted to application of object detection in scene analysis. For example, detected bounding boxes are projected onto multi-scale spatial maps for increasing the accuracy of event recognition~\cite{xiong2015recognize}.

\subsection{Visual recommender systems}
Development of visual recommender systems has become all the more important in the last few years~\cite{you2016picture}. For example, simultaneous image-text co-modeling has resulted in the development of context-aware tweet recommendations~\cite{chen2016context}. Content-based video recommendation system is developed in the paper~\cite{deldjoo2016content}. The VisNet architecture with parallel shallow neural net and VGG16 convolutional neural network (CNN) was fine-tuned like a siamese net, taking input triplets of a query image, a similar image and a negative example~\cite{shankar2017deep}. The clothing, shoes and jewelry from Amazon product dataset are recognized in~\cite{andreeva2018extraction} by an extraction of ResNet-based visual features and a special shallow net. Visual search and recommendations are implemented on Pinterest~\cite{zhai2017visual} using Web-scale object detection and indexing with very deep CNNs~\cite{goodfellow2016deep}.

\section{Materials and methods}
\label{sec:3}

\subsection{Proposed approach}
\label{subsec:3.1}

The main task of this paper is to predict a probability distribution $\pi_c, c \in \{1,...,C\}$, $\sum \limits_{c=1}^{C}{\pi_c}=1$ over $C$ categories as the interest distribution for a new user given a \textit{set} of his or her $M$ photos $X(m), m \in \{1,...,M\}$~\cite{you2016picture}. This distribution can be used either directly to make recommendations or as initial information to solve the cold start problem. We assume that users will grant access to their media files for modeling of their preferences.

In order to extract specific interests from each photo or video, object detection and scene recognition methods were used. At first, we analyzed popular object detection datasets, i.e., MS COCO, OID (Open Image Dataset v4) and ImageNet object datasets, and selected $O=145$ categories from them related to user interests~\cite{grechikhin2019user}. All categories were organized into several high-level groups: outdoors, indoors, sports, food, activity, fashion, musical instruments, transport, services, appliances and toys. In addition, we chose $S=337$ scenes from the united Places365 and Places-Extra69 datasets~\cite{zhou2018places}. 

\begin{figure*}
 \centering
 \includegraphics[width=0.93\linewidth]{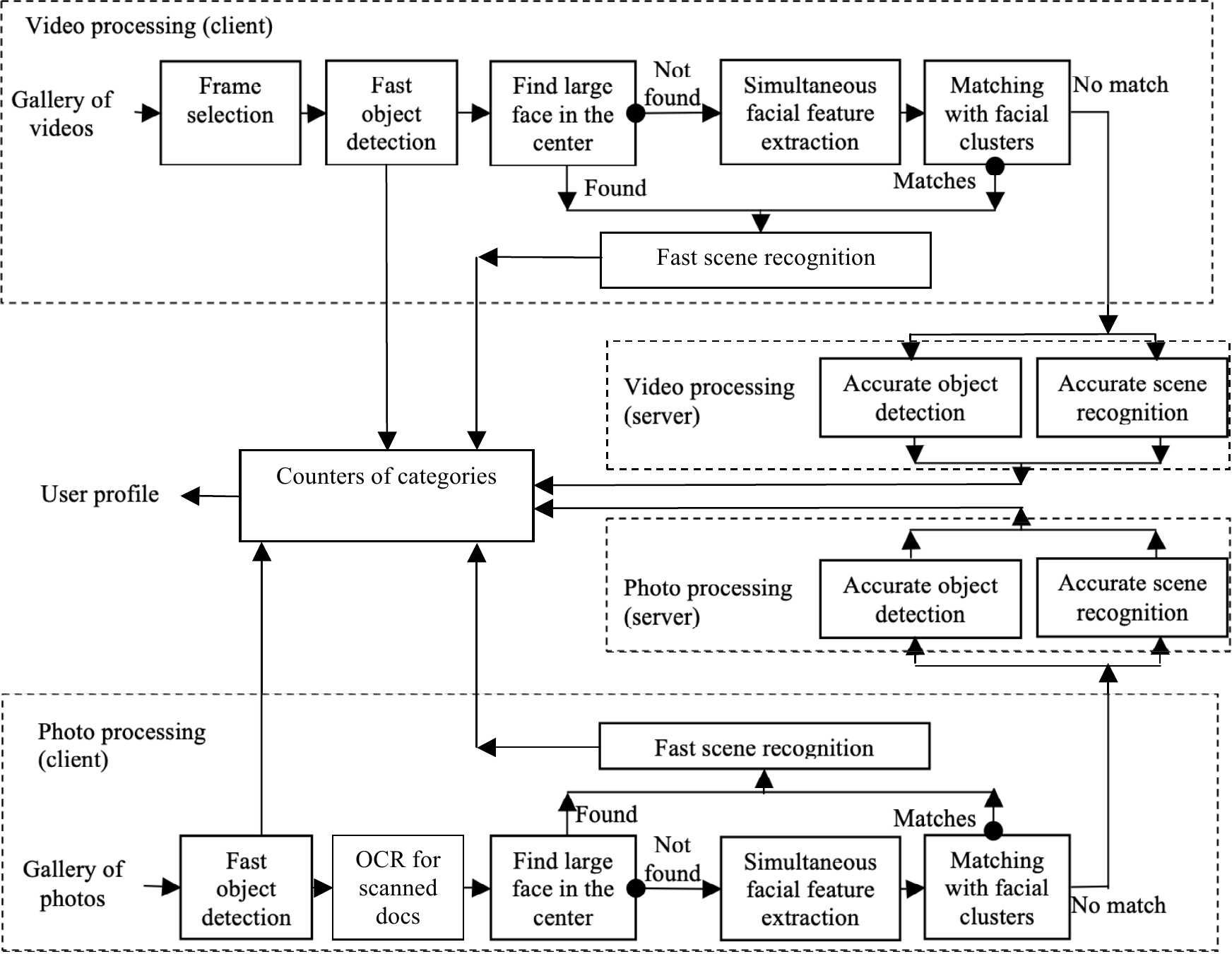}
 \caption{Proposed visual preferences prediction pipeline}
\label{fig:final_pipeline}
\end{figure*}

The proposed pipeline is presented in Fig.~\ref{fig:final_pipeline}. Here, firstly, objects (including faces) on all photos on a mobile device are detected in the ``Fast object detection'' block in an offline mode using efficient CNNs, e.g., MobileNet and SSDLite~\cite{sandler_inverted_2018}. Secondly, it is predicted, whether a photo may be loaded to remote server. At first, we work with scanned documents and detect all text in a photo using existing OCR (optical character recognition) engines, e.g., Firebase Machine learning kit (ML Kit). Next, the detected text is fed into the fully-connected neural network with two hidden layers from our paper~\cite{kopeykina2019automatic}. We demonstrated that it improves the accuracy to more than 97.2\% for specially gathered balanced dataset of sensitive scanned documents. If the whole detected text is classified as sensitive, then this photo is considered as private~\cite{tran2016privacy} and is processed directly on mobile device. In other case we analyze detected facial regions. If the photo is a portrait, i.e., its central part contains at least one face with width greater than a predefined threshold multiplied by a width of the whole photo, then it is also considered as private. In order to estimate this threshold, we gathered 4 galleries of different users with at least 200 portrait images in each gallery and manually labeled all personal photos. After that we computed the minimal width of the face detected by Faster R-CNN with InceptionResNet in a central 2/3 part divided by the width of the photo. As a result, we obtained threshold 0.05 for this ratio, which helps to identify 98.5\% portrait photos.

Thirdly, the following heuristic is used: the photo is considered to be private if it contains a face from rather large facial group (with, e.g., 5 photos made in at least 2 days). Such person is potentially important to the owner of the mobile device so he or she can be supposed to represent the family member or closed friend. In order to apply such an heuristic, all $R \ge 0$ faces detected in all photos are fed into a CNN trained for face identification~\cite{savchenko2017maximum,savchenko2018unconstrained} in order to simultaneously extract numerical feature vectors of faces and predict age and gender~\cite{savchenko2019efficient} in the ``Simultaneous facial feature extraction'' block. As the faces are observed in unconstrained conditions, modern transfer learning and domain adaptation techniques are used for this purpose~\cite{goodfellow2016deep}. According to these methods, the large external dataset of celebrities is used to train a deep CNN. The outputs of one of the last layers of this CNN form $D$-dimensional ($D \gg 1$) embeddings $\mathbf{x}_r = [x_{r;1}, ..., x_{r;D}]$ of the $r$-th facial image from the gallery. These feature vectors are $L_2$-normed in order to provide additional robustness to variability of observation conditions~\cite{cao2018vggface2,savchenko2018efficient}. Fourthly, as the facial images do not contain labels of particular subjects on the photos, the problem of extracting people from the gallery should be solved by clustering methods in the ``Matching with facial clusters'' block. Namely, every face on image should be assigned to one of the labels $1,...,\tilde{R}$, where $\tilde{R}$ is a number of people on images in the user’s gallery~\cite{savchenko2018efficient}. As $\tilde{R}$ is usually unknown, all resulted facial identity feature vectors are grouped by hierarchical agglomerative or density based spatial clustering methods, e.g. DBSCAN. The gender and the birth year of a person in each cluster are estimated by appropriate fusion technique. Selfies are automatically detected using EXIF information about camera model and focal length. An owner of the device is associated with the facial cluster with the largest number of selfies. 
The demography analysis procedure is presented in Algorithm~\ref{algorithm1}. It has two hyper-parameters needed by clustering algorithm, e.g., DBSCAN, namely, the minimal number of faces in a cluster and the maximal distance $\epsilon$ between photos of the same person. The latter is estimated using the datasets suitable for face verification, e.g., Labeled Faces in-the-Wild.

\begin{algorithm}
\caption{Demography analysis in user visual preference prediction} 
\label{algorithm1}
\begin{algorithmic}[1]
\Require gallery of photos ${X(m)}, m \in \{1,...,M\}$
  \State Initialize the number of found faces $R:=0$
  \For{each photo $m \in \{1,...,M\}$}
  \State Feed image $X(m)$ into a fast object detector and detect facial regions
  \For{each detected face}
  \State Assign $R:=R+1$
  \State Feed facial image into the multi-output CNN~\cite{savchenko2019efficient} to extract embeddings $\mathbf{x}_R$ and predict scores of ages $\mathbf{a}_r$, genders $\mathbf{g}_R$ and ethnicities $\mathbf{e}_R$
  \EndFor
  \EndFor
  \State Perform clustering of a set $\{\mathbf{x}_r\},r \in \{1,...,R\}$ to obtain $\tilde{R}$ clusters
  \For{each cluster $\tilde{r} \in \{1,...,\tilde{R}\}$}
  \State Estimate born year, gender and ethnicity by averaging of $\mathbf{a}_r, \mathbf{g}_r, \mathbf{e}_r$ in this cluster
  \EndFor
  \State Compute the histogram $Histo$ of the number of clusters per gender/age range
  \If {there exist only one cluster with maximal number of selfies higher than 0} 
  	\State Obtain cluster $\tilde{r}_{owner} \in \{1,...,\tilde{R}\}$ that contains the maximal number of photos
  	\Return $Histo$ and age, gender and ethnicity of the estimated owner $\tilde{r}_{owner}$ of the mobile device 
  \EndIf
  \Return $Histo$ and the following status: "Photos in the gallery are not enough to perform demography analysis"
\end{algorithmic}
\end{algorithm}

Fifthly, the scenes on the private photos are recognized on the mobile device of a user in the ``Fast scene recognition'' block by using efficient CNNs, e.g. MobileNets. Sixthly, other (``public") photos are sent to remote side (e.g., Flask server) to be processed in the ``Accurate object detection'' block by complex but accurate object detector, e.g., the Faster R-CNN with Inception or InceptionResNet backbone, and scene classifier, e.g., Inception or ResNet CNN, in the ``Accurate scene recognition'' block. After that the photos are immediately removed at the remote server. Seventhly, the detected objects and recognized scenes are mapped on the predefined list of categories and the resulted categories are combined into the user's profile in the ``Counters of categories'' block. The videos are processed similarly: each of 3-5-th frames in each video are selected in the ``Frame selection'' block and the same procedure is repeated, though a whole video is considered as public only if all its frames are marked as public.

\subsection{Representation of visual data using scene recognition and object detection models}
\label{subsec:3.2}

In this subsection we demonstrate how to solve several different complex image recognition tasks with high accuracy using only results of image processing in the pipeline (Fig.~\ref{fig:final_pipeline}). Let us consider details about proposed representation of photos and videos suitable for complex analysis, i.e., scene and event classification. In this case we need for a training set of $N$ users, so that each $n$-th user ($n \in \{1,...,N\}$) is associated with a collection (gallery) of his or her $M_n$ photos $\{X_n(m)\}, m \in \{1,...,M_n\}$. We assume that the preferences of every $n$-th user over all $C$ categories are known.

Nowadays two techniques are mainly applied to represent visual data. The first one is transfer learning~\cite{goodfellow2016deep}, i.e., fine-tuning of pre-trained CNN on the given training set and replacement of the last (logistic regression) layer of the pre-trained CNN to the new layer with Softmax activations and $C$ outputs. If the training sample is rather small to train a deep CNN from scratch, the transfer learning or domain adaptation can be applied~\cite{goodfellow2016deep}. In these methods a large external dataset, e.g. ImageNet-1000, is used to pre-train a deep CNN. As we pay special attention to offline recognition on mobile devices, it is reasonable to use such CNNs as MobileNet v1/v2~\cite{sandler_inverted_2018}. The final step in transfer learning is fine-tuning of this neural network using the training set from the limited sample of instances. This step includes replacement of the last layer of the pre-trained CNN to the new layer with Softmax activations and $C$ outputs. 

This procedure can be modified by replacing the logistic regression in the last layer to more complex classifier. In this case the off-the-shelf features~\cite{savchenko2017maximum} are extracted using the outputs of one of the last layers of pre-trained or fine-tuned CNN. It is especially suitable for small training samples ($N \approx C$) when the results of the fine-tuning is not too accurate. Namely, the images are fed to the CNN, and the outputs of the one-but-last layer are used as the feature vectors. Such deep learning-based feature extractors allow training of a general classifier, e.g., random forest (RF), support vector machine (SVM), multi-layered perceptron (MLP) or gradient boosting that performs nearly as well as if a large training dataset of images from these $C$ classes is available~\cite{goodfellow2016deep}.

As complex photos of events and scenes are characterized by high variability, they are usually composed of parts, some of those parts can be named and correspond to objects~\cite{zhou2018places}. Hence, in this paper we extend such traditional approach with additional features~\cite{rassadin2019scene}. Namely, the $m$-th photo of the $n$-th user $X_n(m)$ is represented with: 
\begin{enumerate}
\item$D$-dimensional vector $\mathbf{f}_n(m)=[f_{n;1}(m),...,f_{n;D}(m)]$ of the off-the-shelf CNN \textit{features (embeddings)} extracted at the penultimate layer of the CNN trained for scene recognition; 
\item \textit{scores} (predictions at the last layer/estimates of scene posterior probability) $\mathbf{p}_n(m)=[p_{n;1}(m),...,p_{n;S}(m)]$ from the last layer of the same CNN ($\sum\limits_{s=1}^{S}{p_{n;s}(m)}=1$).
\end{enumerate}

Moreover, many photos from the one interest category contains identical objects (e.g., ball in the football), which can be detected by contemporary methods, i.e., SSDLite~\cite{sandler_inverted_2018} or Faster R-CNN~\cite{ren2015faster}. These methods detect the positions of several objects in the input image and predict the scores of each class from the predefined set of $O>1$ types. We completely ignore bounding boxes and extract only the sparse vector of confidences $\mathbf{o}_n(m)=[o_{n;1}(m),...,o_{n;O}(m)]$. If there are several objects of the same type, the maximal score is stored in this feature vector.

\begin{algorithm}
\caption{Training of classifiers based on proposed image representation} 
\label{algorithm2}
\begin{algorithmic}[1]
\Require training and validation sets for $C$ classes
\Ensure ensemble of classifiers
  \For{each training image}
  \State Feed the image into a scene CNN and compute the embeddings $\mathbf{f}:=[f_1,...,f_D]$ and scores $\mathbf{p}:=[p_1,...,p_S]$ at the outputs of penultimate and last layers
  \State Feed the image into an object detector and extract vector $\mathbf{o}:=[o_1,...,o_O]$ of maximal confidences for each type of object
  \EndFor
  \State Train classifiers $\mathcal{C}_f,\mathcal{C}_p,\mathcal{C}_o$ using training sets of embeddings, scene scores and detector confidences, respectively
  \For{each validation image}
  \State Extract scene embeddings $\mathbf{f}$ and predict $C$-dimensional confidence scores $\mathbf{cs}_f$ using classifier $\mathcal{C}_f$
  \State Extract scene scores $\mathbf{p}$ and predict $C$-dimensional confidences $\mathbf{cs}_p$ using $\mathcal{C}_p$
  \State Compute maximal confidences of each detected object $\mathbf{o}$ and predict $C$-dimensional confidences $\mathbf{cs}_o$ using $\mathcal{C}_o$
  \EndFor
  \State Assign $\alpha^*:=0$
  \For{all possible weights $w_f, w_s, w_o$}
  \For{each validation image}
	\State Compute confidences $[cs_1,...,cs_C]:=w_f\mathbf{cs}_f+w_p\mathbf{cs}_p+w_p\mathbf{cs}_o$ 
	\State Obtain class with the maximal confidence $c^*:=\underset{c \in \{1,...,C\}}\argmax cs_c$
  \EndFor
  \State Compute accuracy $\alpha$ using predictions $c^*$ of all validation images
  \If {$\alpha^*<\alpha$} 
  \State Assign $\alpha^*:=\alpha, w_f^*:=w_f, w_s^*:=w_s, w_o^*:=w_o$
  \EndIf
  \EndFor
  \Return classifiers $\mathcal{C}_f,\mathcal{C}_p,\mathcal{C}_o$ and their weights $w_f^*, w_s^*, w_o^*$
 \end{algorithmic}
\end{algorithm}

During the recognition process, these three vectors may be combined into a single $(D+S+O)$-dimensional representation, or all of them are classified independently and then the classification results are combined using the simple voting. In this paper we use the latter approach and implemented the classifier fusion technique, which consists of the features and scores from the scene recognition model, scores from the fine-tuned CNN and predictions of the object detection model trained on large dataset. The outputs of individual classifiers are combined with soft aggregation, i.e., the decision is taken in favor of the class with the highest weighted sum of outputs of individual classifier. The weights can be chosen using special validation subset (Algorithm~\ref{algorithm2}). It is important to emphasize that this algorithm does not have hyper-parameters which should be tuned. A new image is classified using steps 7-9, 14-15 of this algorithm, in which $w_f, w_s, w_o$ are replaced by the best weights $w_f^*, w_s^*, w_o^*$.

\subsection{Efficient recognition of a set of photos}
\label{subsec:3.3}

In this subsection we describe how to use such representation for classification of a \textit{set} of photos ${X(m)}, m \in \{1,...,M\}$ of the $n$-th user. For simplicity, we represent the $m$-th photo of the $n$-th user as a $K$-dimensional feature vector $\mathbf{x}_n(m)$, which can be either any of the above-mentioned features (i.e., $K \in \{D, S, O\}$) or their combination (i.e., $K = D+S+O$). Similarly, the $m$-th photo $X(m)$ of the input user is represented with the $K$-dimensional feature vector $\mathbf{x}(m)$. At the second stage, his or her final descriptor $\mathbf{x}$ is produced as a weighted sum of features $\mathbf{x}(m)$, where the weights $w(\mathbf{x}(m))$ may depend on these features. If there is no training data, i.e., $N=0$, then the equal weights will be used, so that conventional averaging with computation of mean feature vector is implemented. However, in this paper we propose to learn the weights $w(\mathbf{x}(m))$ by using a modification of an attention mechanism with a learnable $K$-dimensional vector of weights $\mathbf{q}$~\cite{yang2017neural}. Its authors sequentially combined two attention blocks so that the first aggregated vector is fed into fully-connected layer with $\tanh$ activation and matrix of weights $W \in \mathbb{R}^{K\times K}, \mathbf{b} \in \mathbb{R}^{K\times 1}$ in order to compute $K$-dimensional vector of attention weights $\mathbf{q}^1$. After that the second attention block is used to aggregate the input features. Its usage leads to $K(K+1)$ additional parameters and slower decision-making. In order to improve the run-time complexity, we proposed to reduce the dimensionality of the visual features $\mathbf{x}_n(m)$ by learning the matrix $W_s \in \mathbb{R}^{K\times \tilde{K}}$:
\begin{equation}
\label{eq:squeeze_operator}
 \mathbf{s}(m) = W_s \mathbf{x}(m),
\end{equation}
where $\tilde{K}<K$. The final $\tilde{K}$-dimensional descriptor is computed as follows
\begin{equation}
\label{eq:squeezed_short_attention_so}
 \mathbf{x} =\sum \limits_{m=1}^{M}{w(\mathbf{x}(m))\mathbf{s}(m)},
\end{equation}
where the weights
\begin{equation}
\label{eq:squeezed_weights}
  w(\mathbf{x}(m)) = \frac{\exp(\mathbf{q}^T\mathbf{s}(m))}{\sum \limits_{j=1}^{M}{\exp(\mathbf{q}^T\mathbf{s}(j))}}
\end{equation}
are estimated using the squeezed features (\ref{eq:squeeze_operator}) and an attention mechanism~\cite{yang2017neural}. As a result, the proposed approach (\ref{eq:squeeze_operator})-(\ref{eq:squeezed_weights}) introduces only $(K+1)\cdot \tilde{K}$ parameters to the traditional computation of average features. In such case, we obtain very fast classifier based only on features from scene recognition and object detection models.

Finally, a fully-connected (FC) layer with $C$ multi-label classifiers (logistic regressions) is added to the outputs of attention block (\ref{eq:squeeze_operator}) and the whole model including the $\tilde{K}$-dimensional vector $\mathbf{q}$ is is learned in end-to-end fashion using the available training set. The profile of a user is predicted for the feature vector $\mathbf{x}$ and top-k categories with the highest scores are used to make relevant recommendations.

\subsection{Mobile application}
\label{subsec:3.4}

\begin{figure*}
 \centering
\begin{subfigure}{.3\textwidth}
 \centering
 \includegraphics[width=\linewidth]{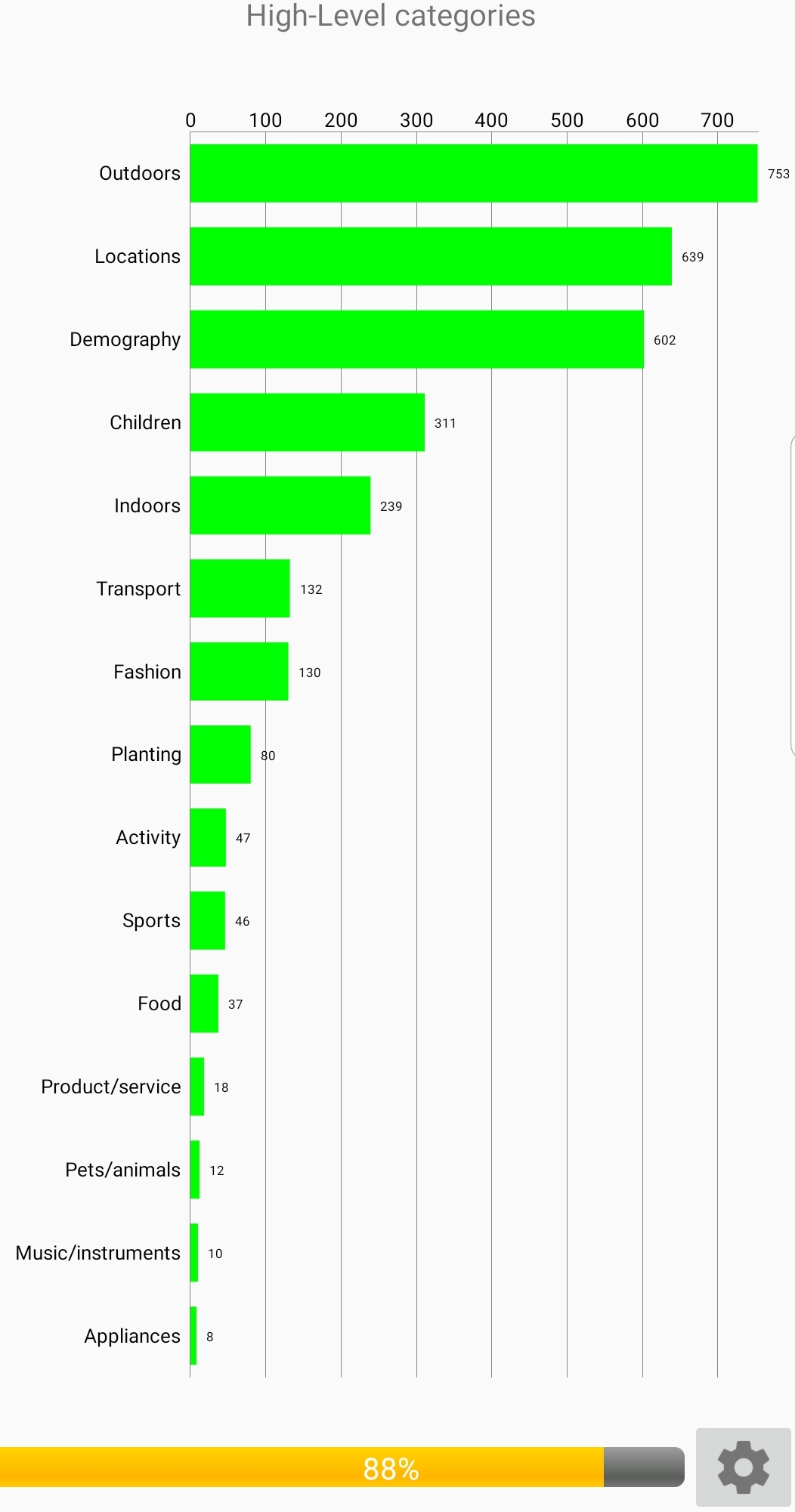}
 \caption{}
 \label{fig:demoUI_sfig1}
\end{subfigure}
\begin{subfigure}{.3\textwidth}
 \centering
 \includegraphics[width=\linewidth]{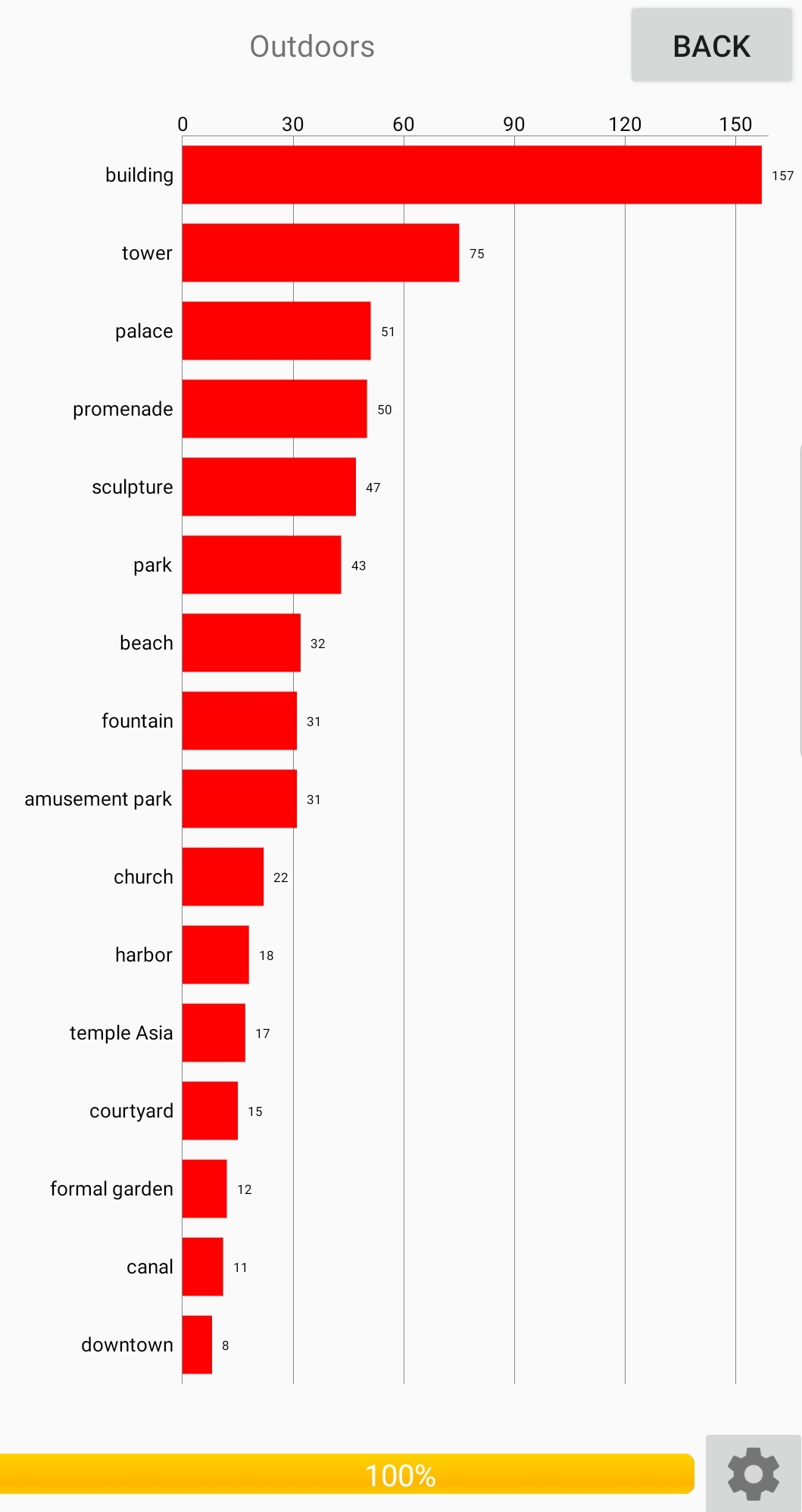}
 \caption{}
 \label{fig:demoUI_sfig2}
\end{subfigure}
\begin{subfigure}{.3\textwidth}
 \centering
 \includegraphics[width=\linewidth]{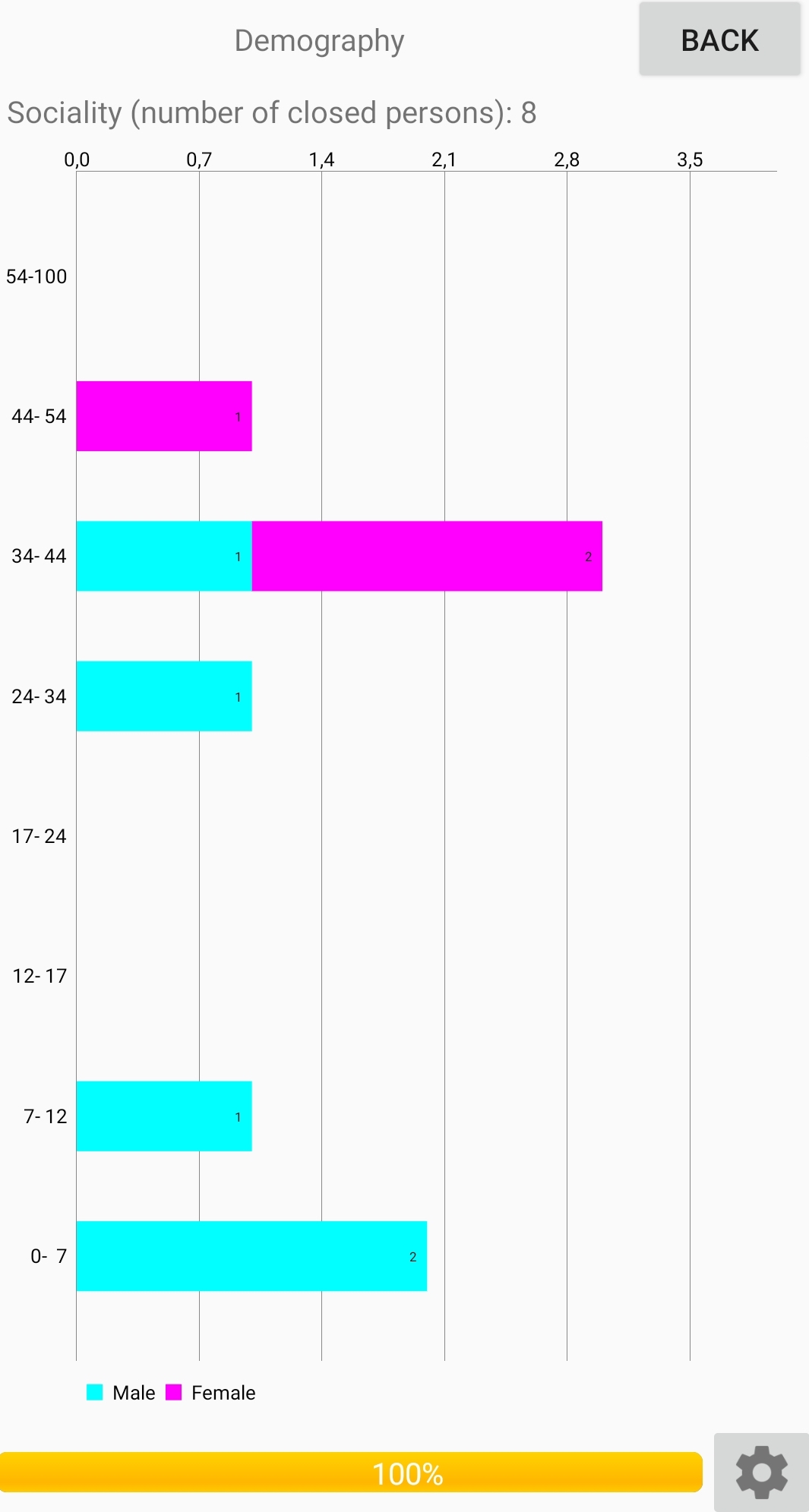}
 \caption{}
 \label{fig:demography_sfig1}
\end{subfigure}
 \caption{User's profile in the mobile demo GUI}
\label{fig:mobile_demo_user_profile}
\end{figure*}

\begin{figure*}
 \centering
\begin{subfigure}{.3\textwidth}
 \centering
 \includegraphics[width=\linewidth]{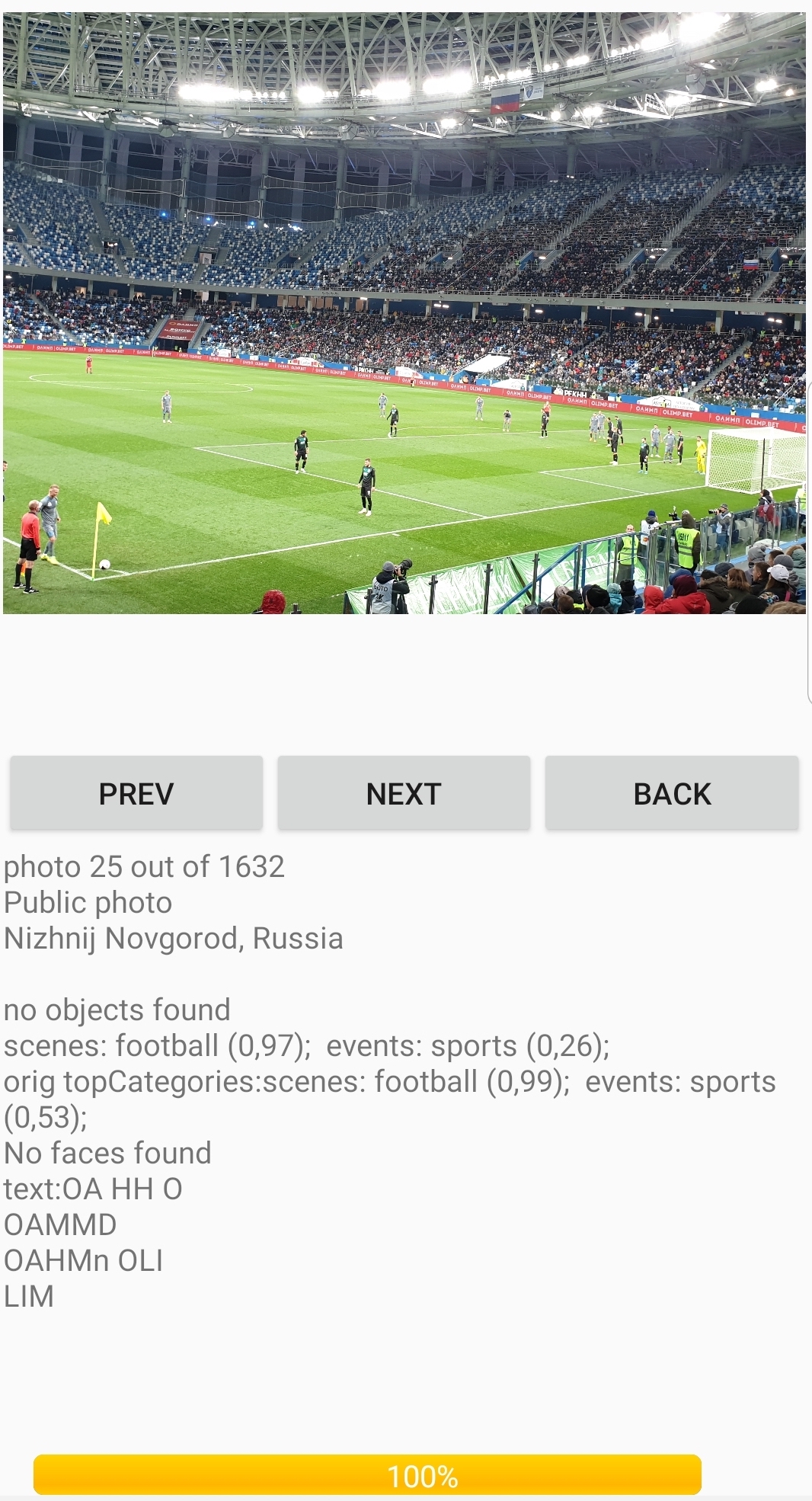}
 \caption{}
 \label{fig:demoUI_sfig3}
\end{subfigure}
\begin{subfigure}{.3\textwidth}
 \centering
 \includegraphics[width=\linewidth]{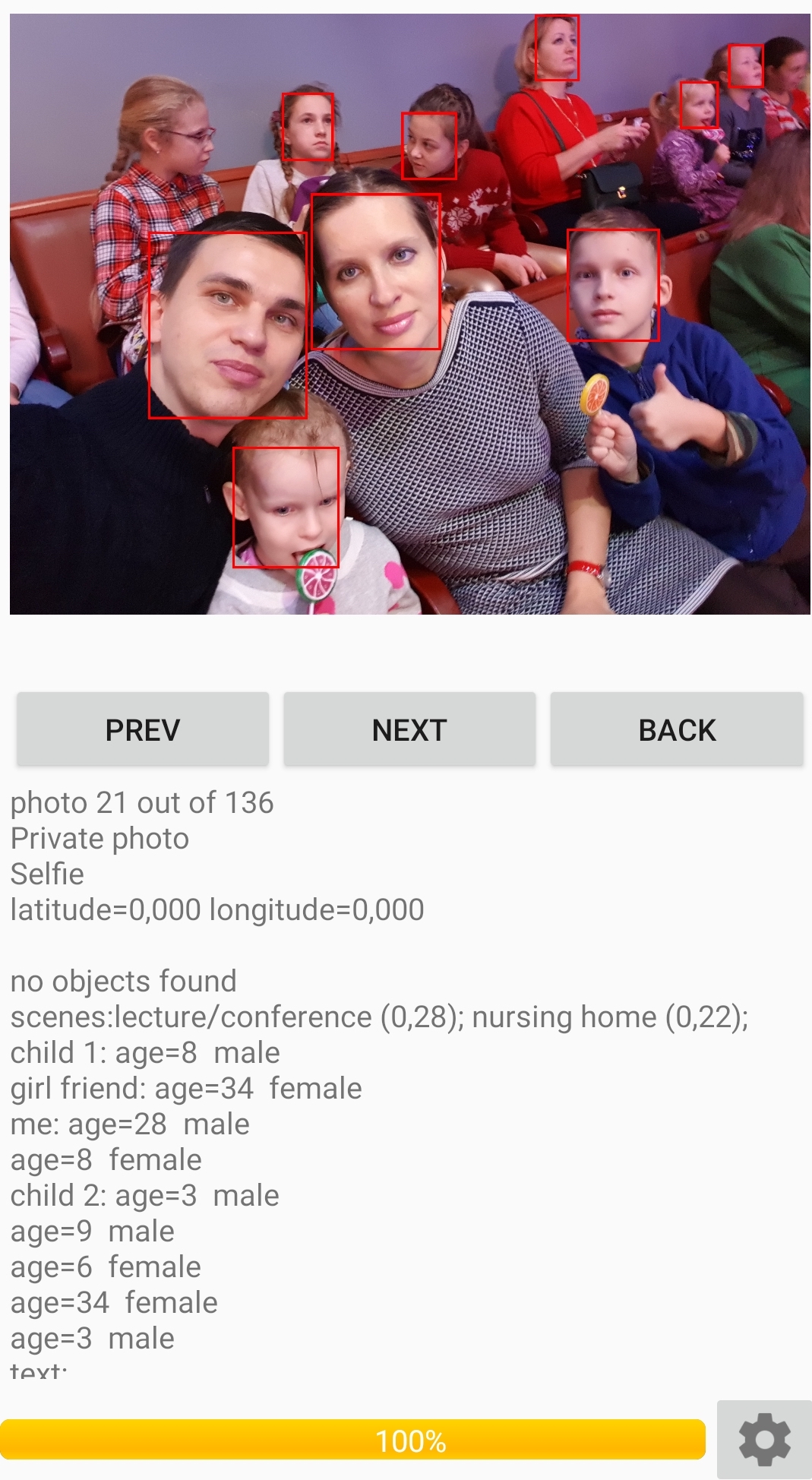}
 \caption{}
 \label{fig:demography_sfig2}
\end{subfigure}
 \caption{Detailed photo analysis in the mobile demo GUI}
\label{fig:mobile_demo_details}
\end{figure*}

We implemented the whole pipeline (Fig.~\ref{fig:final_pipeline}) in the Android demo application\footnote{\url{https://drive.google.com/drive/folders/1rQkJZifq_89pu0sT_UnYXziuxutTpNEN}} (Fig.~\ref{fig:mobile_demo_user_profile}, Fig.~\ref{fig:mobile_demo_details}), which sequentially processes all photos from the gallery in the background thread. The source code of this application is publicly available\footnote{\url{https://github.com/HSE-asavchenko/mobile-visual-preferences}}. The sample screenshot of the main user interface is shown in Fig.~\ref{fig:demoUI_sfig1}. It is possible to tap any bar in this histogram to show a new form with detailed categories (Fig.~\ref{fig:demoUI_sfig2}). We support two special high-level categories: ``Locations'' with most popular cities obtained from geo tags and ``Demography'' with the stacked histograms (Fig.~\ref{fig:demography_sfig1}) of age and gender of the closed persons. If a concrete category is tapped, a ``display'' form appears, which contains a list of all photos from the gallery with this category (Fig.~\ref{fig:demoUI_sfig3}). Tapping the bar in the Demography page (Fig.~\ref{fig:demography_sfig1}) will display the list of all photos of particular subject (Fig.~\ref{fig:demography_sfig2}).

\begin{figure*}[ht!]
 \centering
\begin{subfigure}{.3\textwidth}
 \centering
 \includegraphics[width=\linewidth]{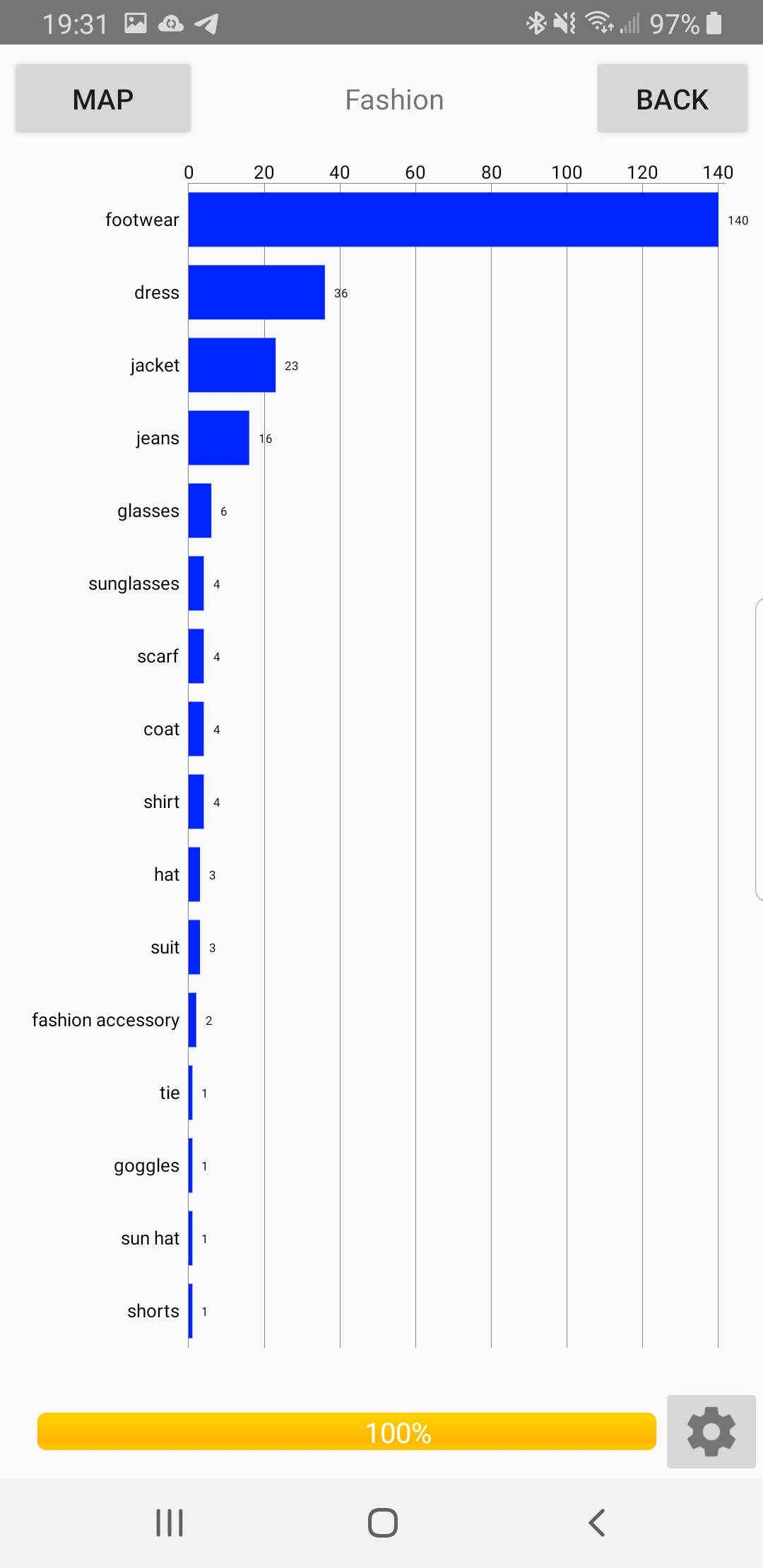}
 \caption{}
 \label{fig:recsys_demo_b}
\end{subfigure}
\begin{subfigure}{.3\textwidth}
 \centering
 \includegraphics[width=\linewidth]{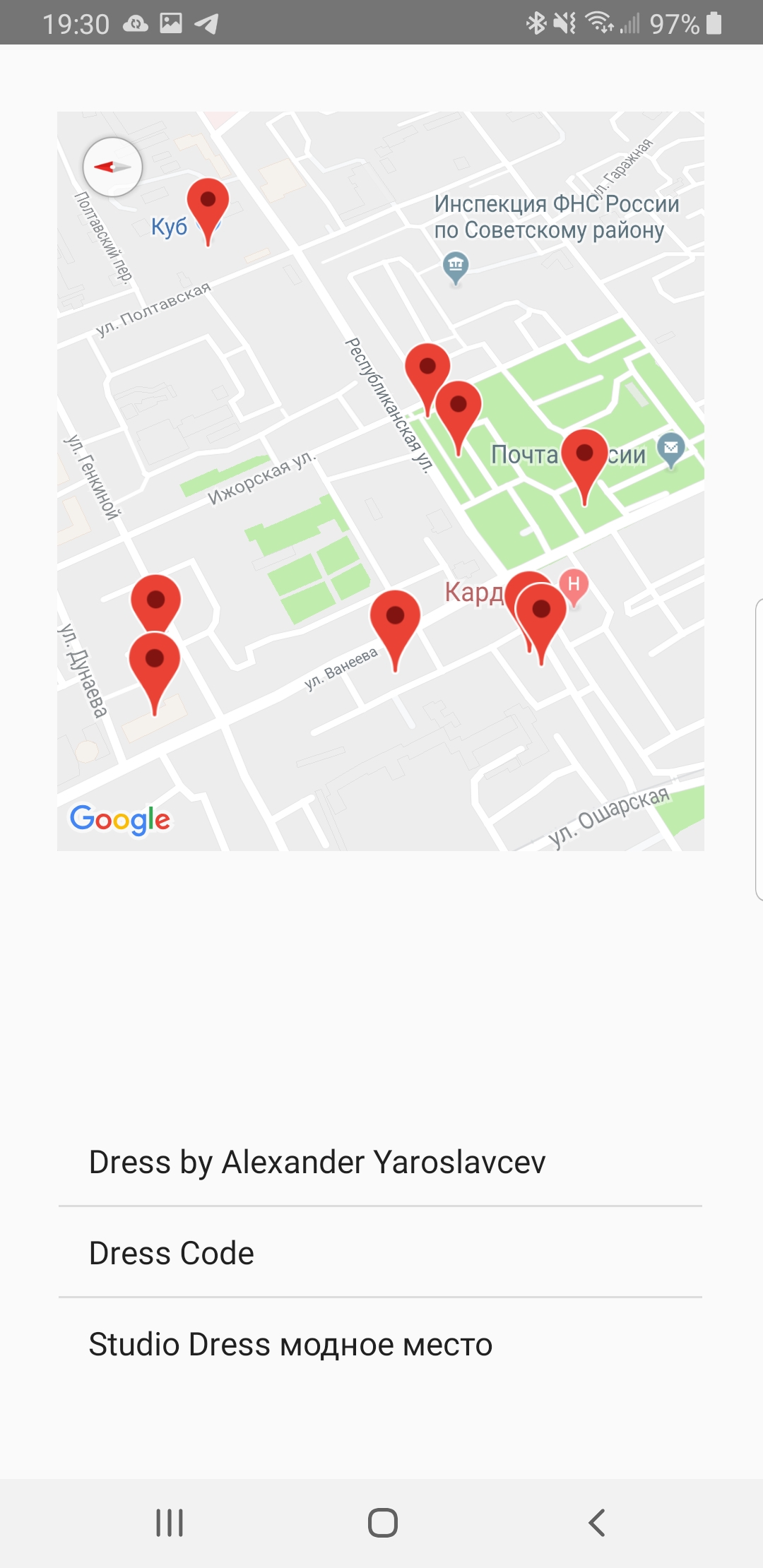}
 \caption{}
 \label{fig:recsys_demo_b}
\end{subfigure}
\begin{subfigure}{.3\textwidth}
 \centering
 \includegraphics[width=\linewidth]{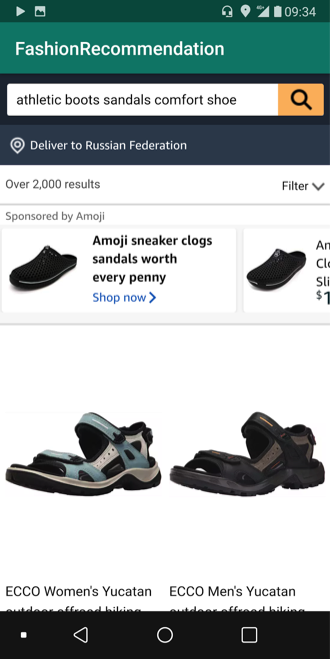}
 \caption{}
 \label{fig:recsys_demo_c}
\end{subfigure}
 \caption{Fashion recommendation: (a) fashion profile; (b) nearby shops; (c) search}
\label{fig:recsys_demo_UI}
\end{figure*}

The gathered profile for each high-level category (Fig.~\ref{fig:demoUI_sfig2}) can be used for recommending locations of shops, restaurants and other places near his or her current geographic location. We added Google maps/Google places and Amazon search functionality so that a request is made to the Google Places API for each identified category and all of the results are consolidated into a single list of store names and markers are placed on an interactive map for each identified store (Fig.~\ref{fig:recsys_demo_UI}). Amazon search for now just opens a search page on the Amazon website with a query for all found categories. 

\section{Experimental study}
\label{sec:4}

\subsection{Scene recognition for user modeling} \label{subsec:4.1a}
In this subsection we explore several known ways to implement fast object detection and scene recognition models used in the proposed pipeline (Fig.~\ref{fig:final_pipeline}). The best scene recognition model was chosen using the above-mentioned subset of the united Places2 dataset. The whole scene dataset was split into the training and test subsets with 8M and 40K images, respectively. This dataset was used to train several CNNs including MobileNet v1/v2~\cite{sandler_inverted_2018} and Inception v3/v4~\cite{szegedy2017inception}. We applied \textit{structural pruning} methods to speed-up offline classification on mobile side. It was found in the preliminary experiments that the best convergence is demonstrated by the pruning with the Taylor expansion~\cite{molchanov2016pruning}. Hence, this method was used to prune approximately 25\% and 35\% of all channels in each convolutional layer. The inference time was measured on MacBook and two Samsung devices (Galaxy Tab S4 tablet PC and Galaxy S9+ phone). The recognition results are presented in Table~\ref{table:scenerec_places_results}. The best values are marked in bold. 

\begin{table*}
\caption{Performance analysis of scene recognition models, subset of Places2 dataset}
\label{table:scenerec_places_results}
\begin{center}
 \begin{tabular}{p{0.08\linewidth}p{0.13\linewidth}p{0.09\linewidth}p{0.1\linewidth}p{0.1\linewidth}p{0.1\linewidth}p{0.1\linewidth}p{0.1\linewidth}}
\hline
& & \multicolumn{3}{c}{MobileNet2 ($\alpha=1.0$)} & \multicolumn{2}{c}{MobileNet2 ($\alpha=1.4$)} & Inception3\\
& & Original & Pruning (25\%) & Pruning (40\%) & Original & Pruning (40\%) & Original \\
\hline
\multicolumn{2}{c}{Top-1 accuracy, \%} & 50.7 & 49.8 & 48.7 & 51.3 & 49.5 & {\bf 53.5} \\
\multicolumn{2}{c}{Top-5 accuracy, \%} & 80.4 & 79.8 & 79.0 & 80.7 & 79.3 & {\bf 83.0}\\
\multicolumn{2}{c}{Precision, \%} & 57.5 & 56.2 & 54.9 & 58.0 & 56.1 & {\bf 60.7}\\
\multicolumn{2}{c}{Recall, \%} & 46.7 & 46.2 & 45.4 & 47.1 & 45.6 & {\bf 48.7}\\
\multicolumn{2}{c}{Size, Mb} & 11.1 & 8.3 & {\bf 6.7} & 20.3 & 12.2 & 91.1 \\
\hline
Inference & MacBook Pro 2015 & 18 & 14 & {\bf 12} & 31 & 29 & 87\\
 time, & Galaxy Tab S4 & 95 & 80 & {\bf 70} & 165 & 140 & 440\\
ms & Galaxy S9+ & 80 & 68 & {\bf 60} & 135 & 110 & 350\\
\hline
 \end{tabular}
\end{center}
\end{table*}

Though the Inception v3 model is the most accurate one, its implementation on mobile device can be inappropriate due to the large running time. Hence, this model is an ideal candidate for the server-side scene recognition only (Fig.~\ref{fig:final_pipeline}). The usage of simplified MobileNet model ($\alpha=1.0$) caused better performance with a very low increase of the error rate. Our implementation of structural pruning improves both running time and memory space, so we decided to use the 25\% pruned MobileNet v2 ($\alpha=1.0$) for scene classification on mobile devices. 

\subsection{Object detection for user modeling} \label{subsec:4.1b}
In this subsection we compared existing detectors using the training/testing dataset described in Section~\ref{sec:3}. By using the balanced training set with no more than 5000 images per category~\cite{grechikhin2019user}, we trained such detectors as Faster R-CNN~\cite{ren2015faster} and SSDLite~\cite{sandler_inverted_2018} using Tensorflow Object Detection API. The testing set contains further 5000 images from each category. Some of the categories are considered as a family of similar categories, e.g. ``animal'' and ``cat'' or ``dog'' categories or ``building'' and ``skyscraper''~\cite{grechikhin2019user}. We took such categories into account while estimating recall (an average rate of detected objects from one class) and mAP (average rate of correctly detected object for particular category to all detected objects of this category). 

\begin{table*}
\caption{Results of object detection testing}
\label{table:objdet_testing}
\begin{center}
 \begin{tabular}{p{0.15\linewidth}p{0.15\linewidth}p{0.15\linewidth}p{0.15\linewidth}p{0.12\linewidth}p{0.1\linewidth}}
\hline
Detector & CNN & Model size, MB & Inference time, s. & Recall, \% & mAP, \% \\
\hline
SSDLite500 & MobileNet v2 & {\bf 23} & {\bf 0.035} & 16.6 & 52.5\\
RetinaNet & ResNet-50 & 51 & 0.432 & 32.9 & 69.2 \\
\multirow{4}{*}{Faster R-CNN} & Inception v3 & 54 & 0.223 & 41.4 & 59.3\\
 & ResNet-50 & 116 & 0.318 & 35.0 & {\bf 63.6} \\
 & ResNet-101 & 189 & 0.847& 44.8 & 53.4 \\
 & InceptionResNet & 251 & 6.57 & {\bf 48.5} & 61.8\\
\hline
 \end{tabular}
\end{center}
\end{table*}

The results including the model size and inference time for one image on MacBook Pro 2015 laptop are shown in Table~\ref{table:objdet_testing}. They are similar to existing studies of object detectors~\cite{huang2017speed}. The highest mAP/recall are obtained by Faster R-CNN with InceptionResNet and ResNet-101 backbones, which are the best candidates for server side processing in our pipeline (Fig.~\ref{fig:final_pipeline}). The choice for the mobile device is more difficult, because the SSDLite models are rather fast (300 ms per photo on Samsung S9+ mobile phone), but their recall is too low due to the small size of most objects from our dataset. The Faster R-CNN is too slow for offline detection on mobile devices (1.2 sec per image on Samsung S9+ for Inception v2 backbone and more than 30 seconds per photo for InceptionResNet).

\subsection{Event recognition} \label{subsec:4.2}
In this subsection we examined the representation of visual data (Subsection~\ref{subsec:3.2}) with scores and embeddings from scene recognition model and score from object detection model in event recognition problem. Two datasets were explored, namely, 1) PEC (Photo Event Collection)~\cite{bossard2013event} with 61,000 images from 807 collections of 14 social event classes (birthday, wedding, graduation, etc.); and 2) WIDER (Web Image Dataset for Event Recognition)~\cite{xiong2015recognize} with 50,574 images and 61 event categories (parade, dancing, meeting, press conference, etc).

We used standard train/test split for both datasets proposed by their creators. In the PEC dataset we directly assigned the collection-level label to each image contained in this collection and simply use the image itself for event recognition, without any meta information such as temporal information similarly to recent paper~\cite{wang2018transferring}. The results of client-side models (MobileNet scene model and SSDLite object detector) and server-side (Inception scene model and Faster R-CNN detector with InceptionResNet backbone) are shown in Table~\ref{table:pec_mobilenet} and Table~\ref{table:wider_mobilenet} for the PEC and WIDER, respectively. In addition to client-side and server-side models, we used the whole pipeline (Fig.~\ref{fig:final_pipeline}) here. It automatically selects private photos and recognizes them with the client-side models. All other photos are processed by the server-side models.

\begin{table*}
\begin{center}
\caption{Event recognition results, Photo Event Collection dataset}
\label{table:pec_mobilenet}
\begin{tabular}{lll}
\noalign{\smallskip}
\hline
Features & Classifier & Accuracy, \% \\
\noalign{\smallskip}
\hline
\noalign{\smallskip}
\multirow{3}{*}{MobileNet, scores} & Random Forest & 55.76 \\
 & Linear SVM & 51.66 \\
 & Fine-tuned & 61.11 \\ \hline
\multirow{3}{*}{MobileNet, embeddings} & Random Forest & 57.25 \\
 & Linear SVM & 59.72 \\
 & Fine-tuned & 62.13 \\ \hline
\multirow{3}{*}{SSD + MobileNet} & Random Forest & 30.84 \\
 & Linear SVM & 42.18 \\
 & Fine-tuned (new FC layer) & 40.16 \\ \hline
\multirow{3}{*}{Proposed representation (client-side classifiers)}& Random Forest & 57.60 \\
 & Linear SVM & 60.92 \\
 & Fine-tuned & 63.34 \\ \hline \hline
\multirow{3}{*}{Inception v3, scores} & Random Forest & 57.15 \\
 & Linear SVM & 52.64 \\
 & Fine-tuned & 61.81 \\ \hline
\multirow{3}{*}{Inception v3, embeddings} & Random Forest & 57.50 \\
 & Linear SVM & 61.82 \\
 & Fine-tuned & 63.68 \\ \hline
\multirow{3}{*}{Faster R-CNN + InceptionResNet} & Random Forest & 43.32 \\
 & Linear SVM & 48.83 \\
 & Fine-tuned (new FC layer) & 47.45 \\ \hline
\multirow{3}{*}{Proposed representation (server-side classifiers)}& Random Forest & 59.10 \\
 & Linear SVM& 62.87 \\
& Fine-tuned & 64.98 \\
\hline\hline
\multirow{3}{*}{Complete pipeline} & Random Forest & 58.25 \\
 & Linear SVM & 61.84 \\
 & Fine-tuned & 63.95 \\
 \hline
\end{tabular}
\end{center}
\end{table*}

\begin{table*}
\begin{center}
\caption{Event recognition results, WIDER dataset}
\label{table:wider_mobilenet}
\begin{tabular}{lll}
\noalign{\smallskip}
\hline
Features & Classifier & Accuracy, \% \\
\noalign{\smallskip}
\hline
\noalign{\smallskip}
\multirow{3}{*}{MobileNet, scores} & Random Forest & 40.18 \\
 & Linear SVM & 36.29 \\
 & Fine-tuned & 40.49 \\ \hline
\multirow{3}{*}{MobileNet, embeddings} & Random Forest & 42.32 \\
 & Linear SVM & 48.31 \\
 & Fine-tuned & 49.48 \\ \hline
\multirow{3}{*}{SSD + MobileNet} & Random Forest & 14.69 \\
 & Linear SVM & 19.91 \\
 & Fine-tuned & 12.91 \\ \hline
\multirow{3}{*}{Proposed representation (client-side classifiers)}& Random Forest & 43.55 \\
 & Linear SVM & 48.91 \\
 & Fine-tuned & 49.80 \\ \hline \hline
\multirow{3}{*}{Inception v3, scores} & Random Forest & 41.73 \\
 & Linear SVM & 35.91 \\
 & Fine-tuned & 41.66 \\ \hline \hline
\multirow{3}{*}{Inception v3, embeddings} & Random Forest & 42.57 \\
 & Linear SVM & 50.47 \\
 & Fine-tuned & 50.96 \\ \hline
\multirow{3}{*}{Faster R-CNN + InceptionResNet} & Random Forest & 27.23 \\
 & Linear SVM & 28.66 \\
 & Fine-tuned (new FC layer) & 21.27 \\ \hline
\multirow{3}{*}{Proposed representation (server-side classifiers)}& Random Forest & 45.92 \\
 & Linear SVM & 51.59 \\
& Fine-tuned & 51.76 \\
\hline\hline
\multirow{3}{*}{Complete pipeline} & Random Forest & 45.08 \\
 & Linear SVM & 50.94 \\
 & Fine-tuned & 51.17 \\
\hline
\end{tabular}
\end{center}
\end{table*}

If the fine-tuned CNNs are used in an ensemble, our lightweight MobileNet-based models improved the previous state-of-the-art for PEC from 62.2\%~\cite{wang2018transferring} even for the client-side model (63.34\%). Our server-side model is even better (accuracy 64.98\%). If the fine-tuned CNNs are used in an ensemble, our lightweight MobileNet-based models improved the previous state-of-the-art for the PEC dataset from 62.2\% to 63.34\%. More importantly, we over-perform the known state-of-the-art by using linear SVM for features and score extracted by pre-trained models from previous subsections.

If only pre-trained scene recognition and object detection models are used in the proposed approach in order to prevent additional inference in fine-tuned models, the resulted accuracy for WIDER dataset is only 0.17-0.89\% lower. The accuracy of our MobileNet-based classifiers for this dataset is 7.4-9.3\% higher when compared to the best results (42.4\%) from original paper~\cite{xiong2015recognize}. In contrast to the PEC dataset, WIDER does not contain many facial images of the same subjects and scanned sensitive documents. As a result, most images are considered to be public and the overall accuracy of the complete pipeline is very close to the server-side models.

\subsection{Recognition of a set of images based on user's profile} \label{subsec:4.3}

In the next experiment the proposed aggregation of image features into a single user descriptor (\ref{eq:squeeze_operator})-(\ref{eq:squeezed_weights}) for our representation of images is studied. As there is no publicly available labeled datasets of photo galleries for different users, we used the Amazon Fashion~\cite{kang2017visually} dataset that contains $500000$ entries of $N=16000$ unique users interacting with $40000$ products from $D=75$ fashion categories. There is a single unique item on each picture that belongs to one or more classes. The number $M_n$ of items purchased by a user varies from $5$ to $40$; an average user has interacted with $8$ unique items. The images are grouped by user, and each user is associated with a $C$-dimensional target vector, so that the $c$-th component is equal to 1 if the user has interacted with at least one item from the $c$-th category, and 0, otherwise. The united feature vector obtained by merging the scores and embeddings of scene recognition MobileNet model with the output of SSDLite object detector was used.

We implemented the following aggregation techniques: 1) average pooling of fine-tuned features; 2) pooling of all features with one attention block (\ref{eq:squeezed_short_attention_so}) and additional context gating (CG), which applies a scaling mask to the resulting aggregated vector~\cite{miech2017learnable}; 3) proposed usage of 1-layer attention with reduced ($~\tilde{K}=128$) features (\ref{eq:squeeze_operator})-(\ref{eq:squeezed_weights}); and 4) proposed usage of reduced ($~\tilde{K}=128$) features with two attention blocks recommended in the original paper~\cite{yang2017neural}. The resulted descriptor were fed into FC layer in order to obtain the final profile as described in the last paragraph of Subsection~\ref{subsec:3.3}. The aggregation models were trained on the 70\% of the dataset. The algorithms were tested on the for remaining 30\% users in order to predict $C$ interest probabilities, among which top $k$ interests were recommended. The model size (excluding the size of the feature extractor) and the dependence of precision, recall, F1-score and AUC (area under the ROC curve) on the number $k$ of top categories (with the highest probabilities) for each aggregation technique and the best configuration are shown in Table~\ref{table:aggregation_fashion}. 

\begin{table*}
\begin{center}
\caption{Comparison of aggregation techniques, Amazon Fashion dataset}
\label{table:aggregation_fashion}
\begin{tabular}{ccp{0.15\linewidth}p{0.2\linewidth}p{0.22\linewidth}}
\hline
Metric& Average& 1-Attention + CG & Proposed 1-Attention (\ref{eq:squeeze_operator})-(\ref{eq:squeezed_weights}) & Proposed 2-Layers Attention + FC\\
\hline
Size, MB& {\bf 1.3} & 4.5 & 2.18 & 3.5 \\
\hline
Decision time, ms& {\bf 40} & 43 & 41 & 47 \\
\hline
Precision@k=5& 0.29 & 0.44 & {\bf 0.51} & 0.50 \\
Recall@k=5& 0.56 & 0.61 & 0.66 & {\bf 0.68}\\
F1-score@k=5& 0.38& 0.51& {\bf 0.58}& {\bf 0.58}\\
AUC@k=5& 0.50& 0.60& {\bf 0.62}& 0.61\\
\hline
Precision@k=10& 0.18 & 0.43 & {\bf 0.51} & 0.50 \\
Recall@k=10& 0.70 & 0.63 & 0.67 & {\bf 0.68}\\
F1-score@k=10& 0.29& 0.51& {\bf 0.58}& {\bf 0.58}\\
AUC@k=10& 0.52& 0.51& {\bf 0.71}& {\bf 0.71}\\
\hline
 \end{tabular}
\end{center}
\end{table*}

Here the F1-score of traditional feature aggregation is 13-31\% lower when compared to learnable pooling techniques. The proposed approach (\ref{eq:squeeze_operator})-(\ref{eq:squeezed_weights}) with one attention layer is the most appropriate for mobile applications due to the lowest number of parameters and quality metrics higher than almost all other aggregation methods.

In the next experiment we examined event recognition task with the PEC dataset~\cite{bossard2013event}, in which a \textit{set} of images from an album is observed. We used conventional split into the training set with 667 albums and testing set with 140 albums. We used two techniques in order to obtain a final descriptor of a set of images, namely, 1) simple averaging of features of individual images in a set; and 2) proposed implementation of neural attention mechanism (\ref{eq:squeeze_operator})-(\ref{eq:squeezed_weights}) for $L_2$-normed features. In the latter case the attention weights are learned using the sets with 10 randomly chosen images from all albums in order to make identical shape of input tensors. As a result, 667 training subsets with 10 images were obtained. The recognition accuracies are presented in Table~\ref{table:pec_attention}. Here we provided the best-known results for this dataset~\cite{wang2017recognizing,wu2015learning}.

\begin{table*}
\begin{center}
\caption{Event recognition in a set of images (album), PEC dataset}
\label{table:pec_attention}
\begin{tabular}{cccc}
\hline
Features & Aggregation & Accuracy, \%\\
\hline
\multirow{2}{*}{MobileNet v2 ($\alpha=1.0$)+SSDLite}& Average & 86.42 \\
 & Proposed 1-Attention (\ref{eq:squeeze_operator})-(\ref{eq:squeezed_weights}) & {\bf 89.29} \\
\multirow{2}{*}{MobileNet v2 ($\alpha=1.4$)+SSDLite}& Average & 86.44 \\
& Proposed 1-Attention (\ref{eq:squeeze_operator})-(\ref{eq:squeezed_weights}) & 87.36 \\
\multirow{2}{*}{Inception v3+Faster R-CNN} & Average & 86.43\\
 & Proposed 1-Attention (\ref{eq:squeeze_operator})-(\ref{eq:squeezed_weights}) & 87.86 \\
\hline
\multirow{2}{*}{AlexNet} & CNN-LSTM-Iterative~\cite{wang2017recognizing}& 84.5 \\
 & Aggregation of representative features~\cite{wu2015learning} & 87.9\\
\multirow{2}{*}{ResNet-101} & CNN-LSTM-Iterative~\cite{wang2017recognizing}& 84.5 \\
 & Aggregation of representative features~\cite{wu2015learning} & 89.1 \\
\hline
 \end{tabular}
\end{center}
\end{table*}

In all cases attention-based aggregation is 1-3\% more accurate when compared to classification of average features. As one can notice, the proposed implementation of attention mechanism achieves the state-of-the-art results, though we used much faster CNNs (MobileNet and Inception rather than AlexNet and ResNet-101). The most remarkable fact here is that the best results are achieved for the most simple model (MobileNet v2 $\alpha=1.0$), which can be explained by the lack of training data. What is more important, we do not need to fine-tune existing scene recognition models, so the implementation of event recognition in an album will be very fast.

\subsection{Examples} \label{subsec:4.4}

\begin{figure*}[h!]
 \centering
\begin{subfigure}{1\textwidth}
 \centering
 \includegraphics[width=.6\linewidth]{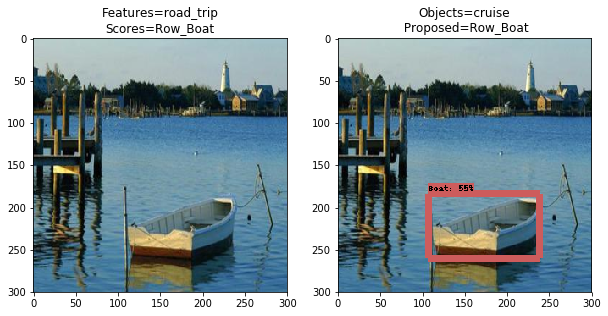}
 \caption{}
 \end{subfigure}

\begin{subfigure}{1\textwidth}
 \centering
 \includegraphics[width=.6\linewidth]{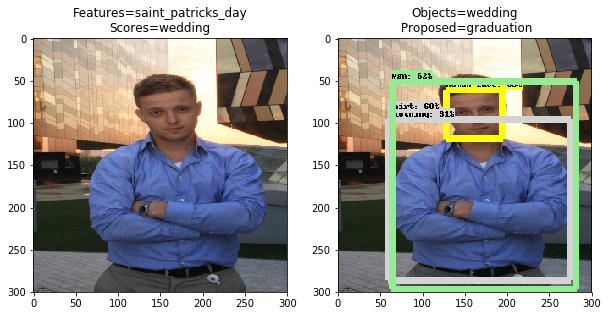}
 \caption{}
\end{subfigure}

\begin{subfigure}{1\textwidth}
 \centering
 \includegraphics[width=.6\linewidth]{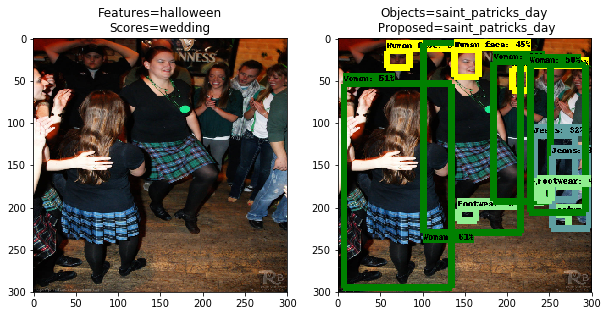}
 \caption{}
\end{subfigure}
 \caption{Sample results of event recognition }
\label{fig:event_recognition_sample}
\end{figure*}

In the last subsection we provide several qualitative examples for the usage of our pipeline. Firstly, the results of correct event recognition using the proposed ensemble (Algorithm~\ref{algorithm2}) are presented on Fig.~\ref{fig:event_recognition_sample}. Here the title of each left image displays the result of event recognition using only features $\mathbf{x}_n$ or scores $\mathbf{p}_n$. Each right image shows the object detection results. Its title contains the event prediction based on object scores and our ensemble. As one can notice, the proposed representation manage to obtain reliable solution even when individual classifiers make wrong decisions.

\begin{figure*}[ht!]
 \centering
 \begin{subfigure}{0.49\textwidth}
  \includegraphics[height=0.1\textheight]{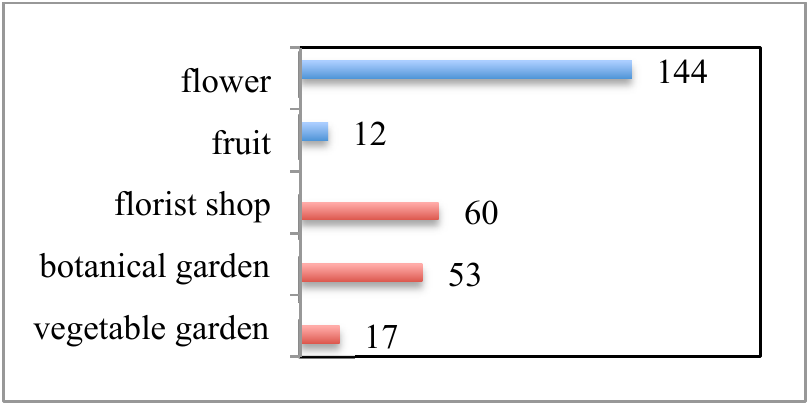}
  \caption{}
 \end{subfigure}
 \begin{subfigure}{0.49\textwidth}
  \includegraphics[height=0.13\textheight]{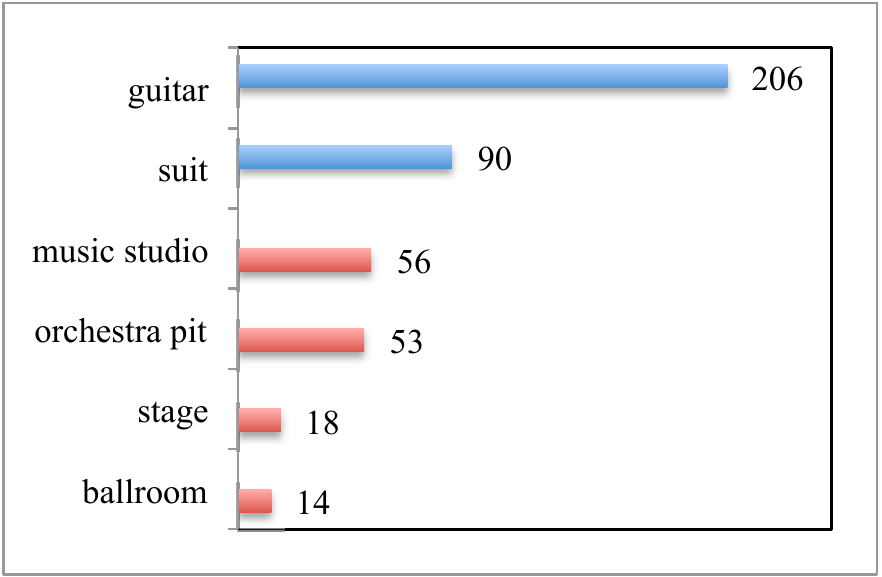}
  \caption{}
 \end{subfigure}
 
 \begin{subfigure}{0.49\textwidth}
  \includegraphics[width=\textwidth, height=0.21\textheight]{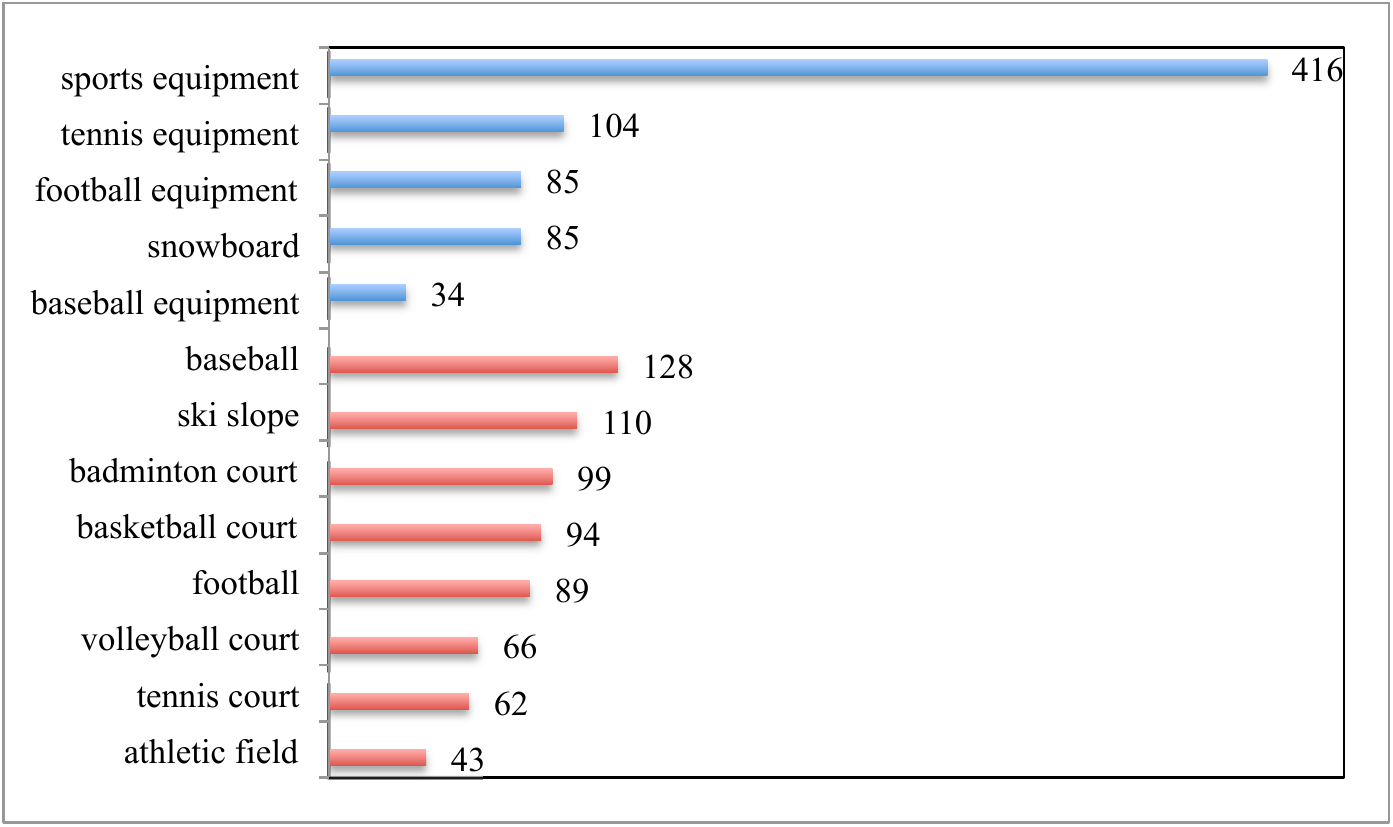}
  \caption{}
 \end{subfigure}
 \begin{subfigure}{0.49\textwidth}
  \includegraphics[height=0.21\textheight]{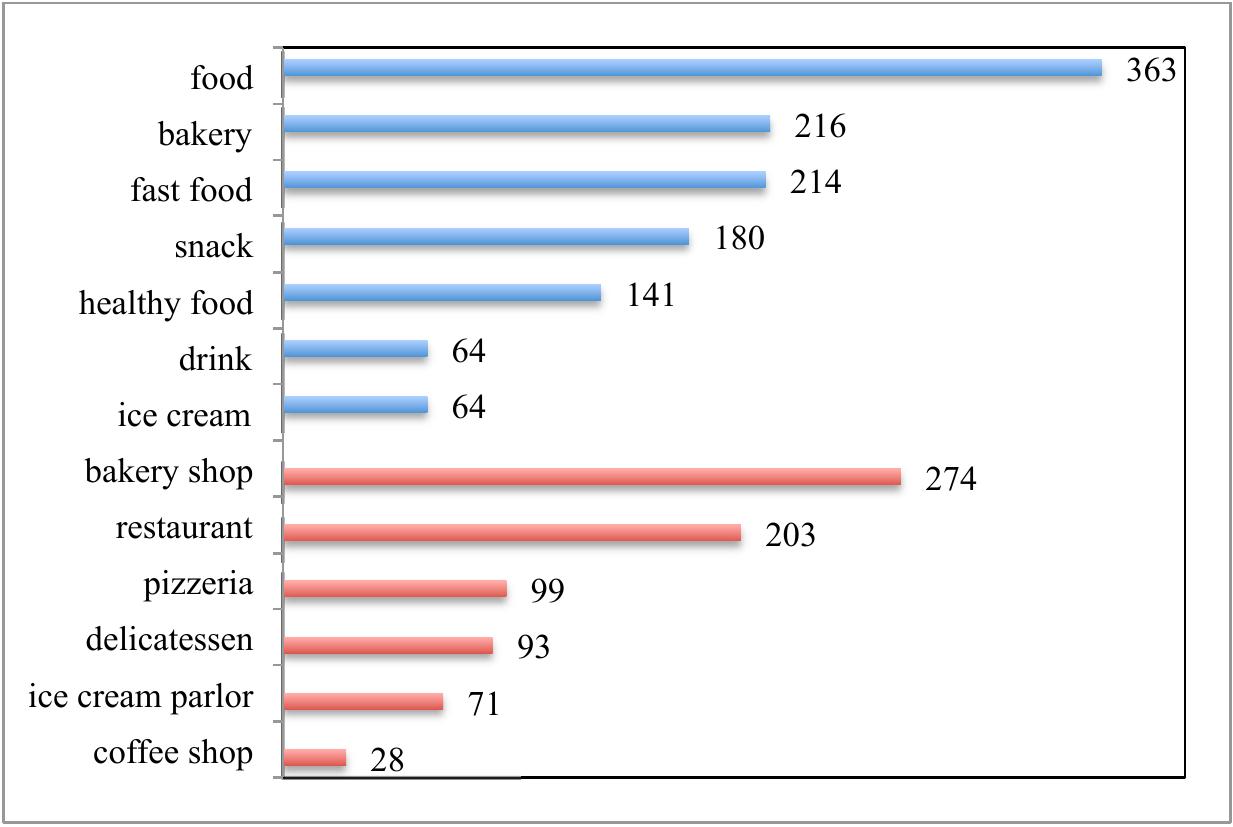}
  \caption{}
 \end{subfigure}
 
 \begin{subfigure}{0.49\textwidth}
    \includegraphics[height=0.17\textheight]{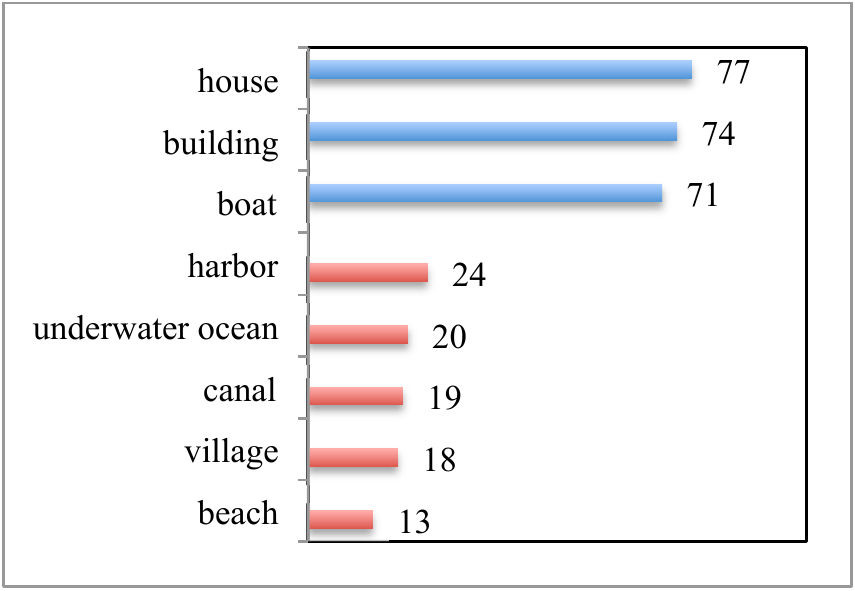}
    \caption{}
  \end{subfigure}
  \begin{subfigure}{0.49\textwidth}
    \includegraphics[height=0.17\textheight]{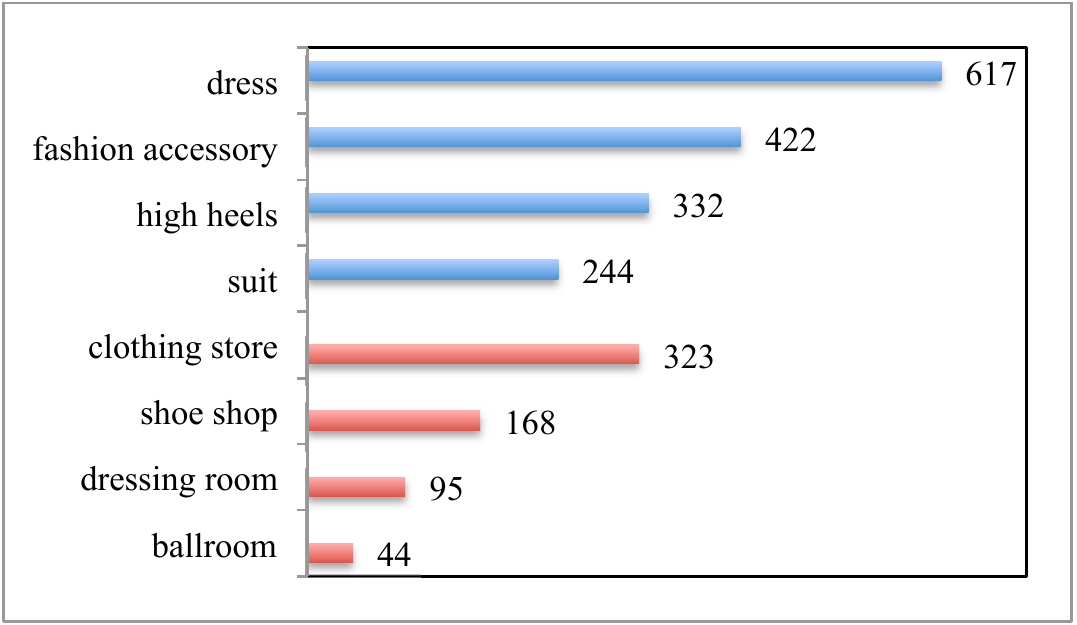}
    \caption{}
  \end{subfigure}
  \caption{Profiles gathered for images from the web using client-side models: (a) Nature; (b) Music; (c) Sport; (d) Food; (e) Traveling; (f) Fashion.}
  \label{fig:web_photos}
\end{figure*}

Secondly, we present several profiles obtained with the proposed approach (Fig.~\ref{fig:final_pipeline}) by processing real galleries of photos. At first, we took six keywords and automatically retrieved images from the Web by feeding these keywords into the visual search engine. Fig~\ref{fig:web_photos} contains the scene (red bars) and object (blue bars) categories for the top predictions of user interests by lightweight models. Here the set of images gathered by keyword is processed very accurately. Though several detected objects are still too general, e.g., building/house for traveling and suit for music, the red bar of top scenes (Fig.~\ref{fig:web_photos}) contains the categories which characterize the keyword perfectly well.

\begin{figure*}[ht!]
 \centering
  \begin{subfigure}{0.64\textwidth}
    \includegraphics[height=0.33\textheight]{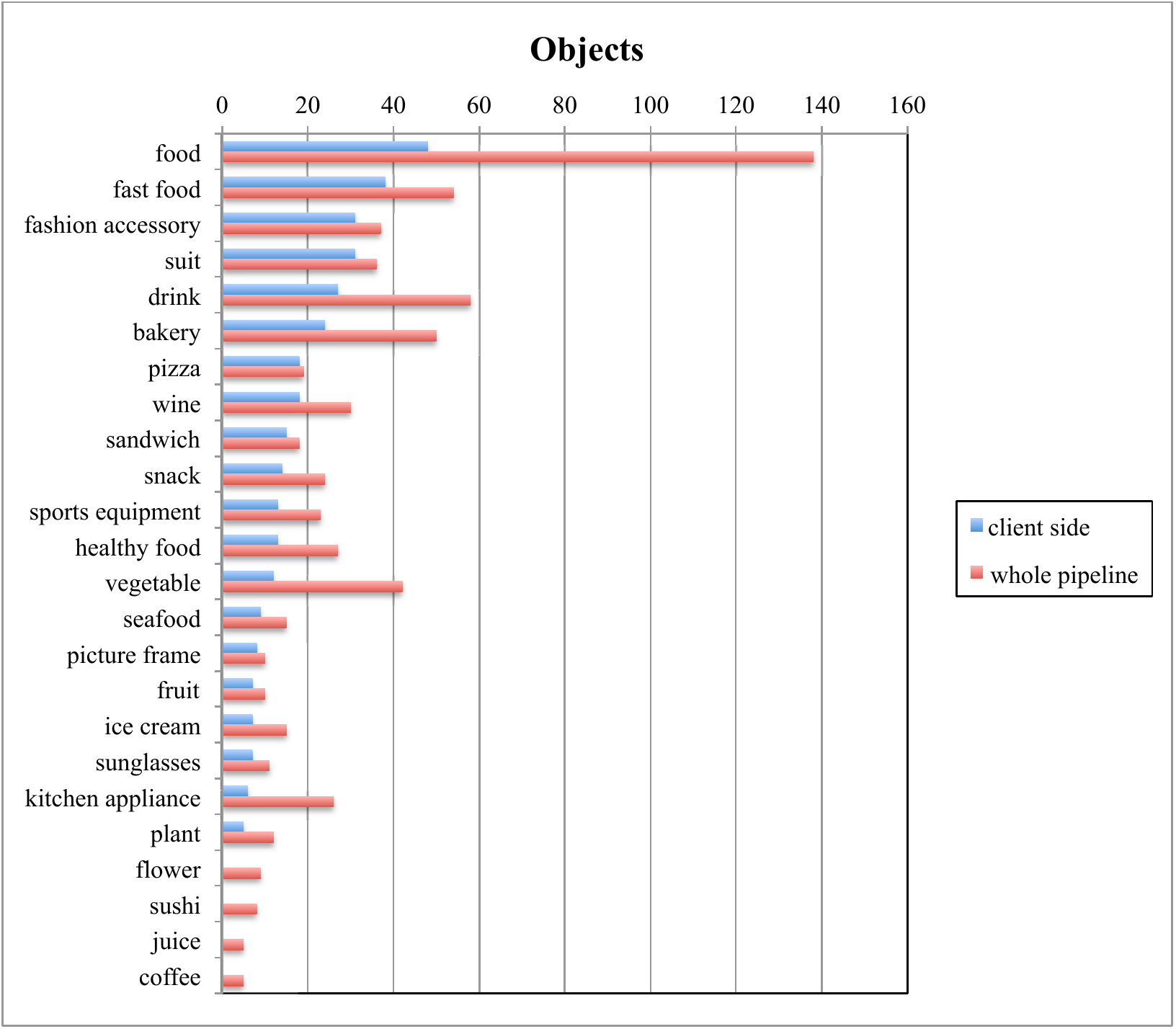}
    \caption{}
  \end{subfigure}
  \begin{subfigure}{0.34\textwidth}
    \includegraphics[height=0.2\textheight]{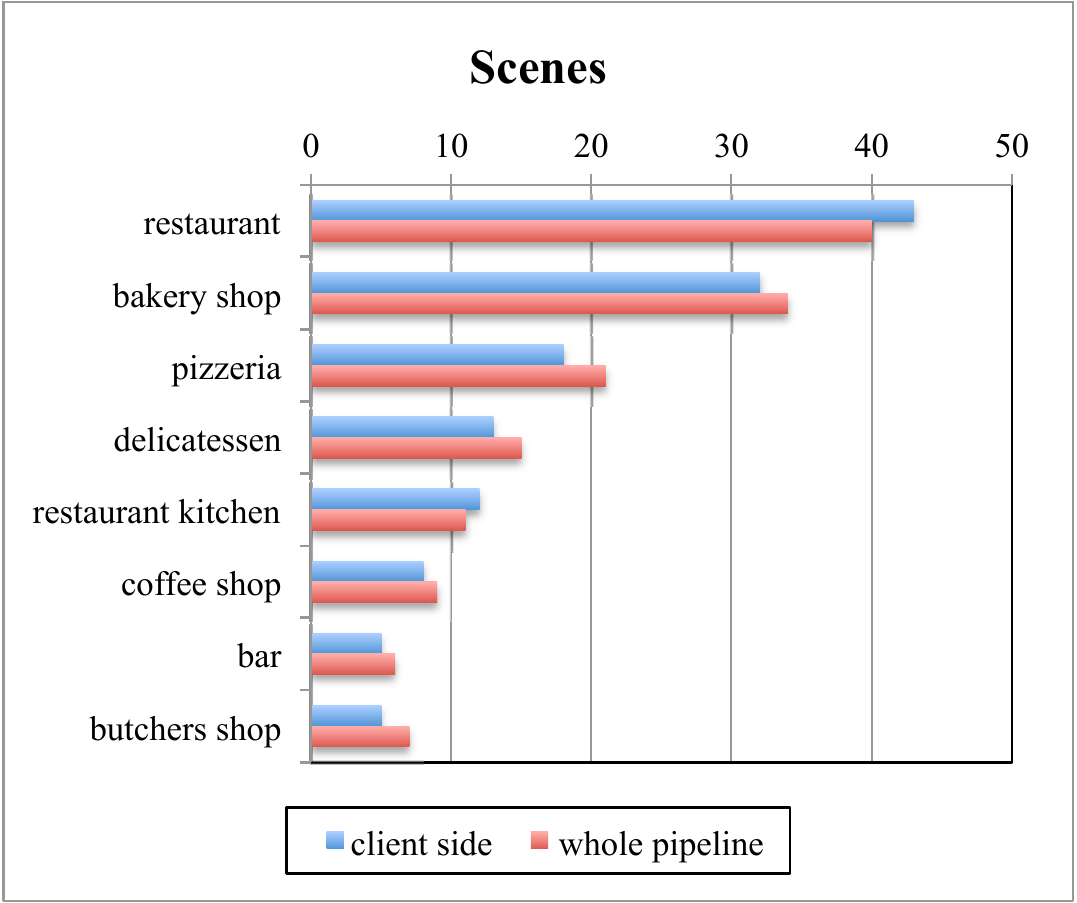}
    \caption{}
  \end{subfigure}
  
  \begin{subfigure}{0.64\textwidth}
    \includegraphics[height=0.33\textheight]{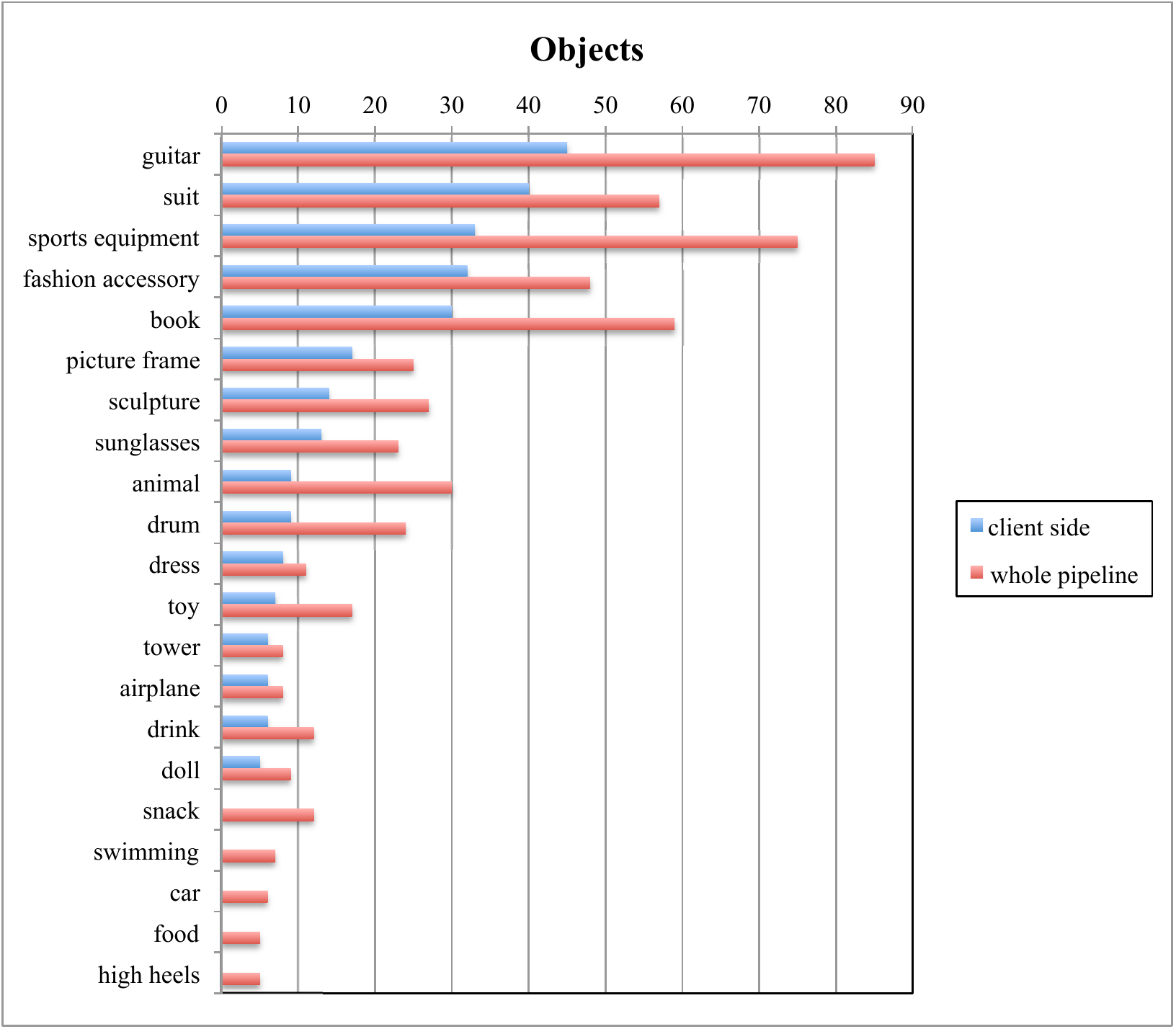}
    \caption{}
  \end{subfigure}
  \begin{subfigure}{0.34\textwidth}
    \includegraphics[height=0.2\textheight]{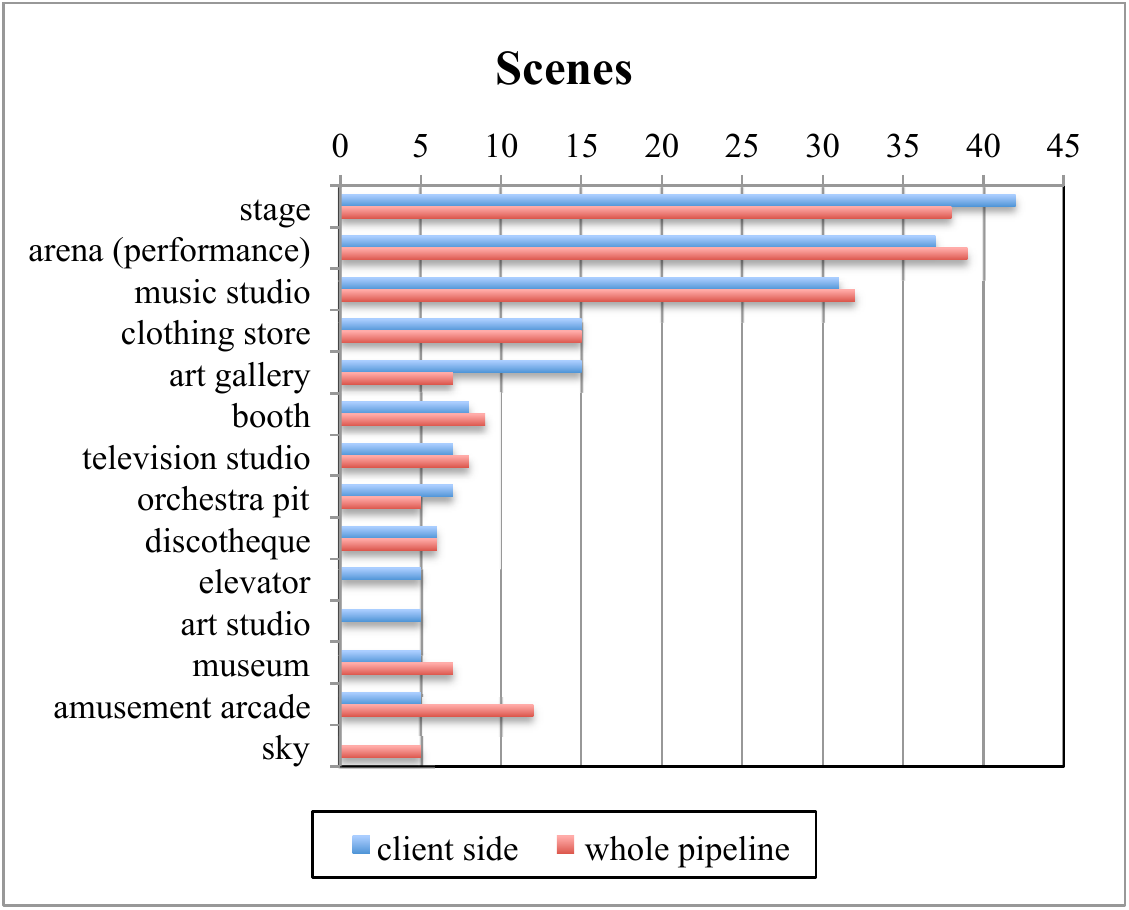}
    \caption{}
  \end{subfigure}
  \caption{Profiles gathered for Instagram photos. (a)-(b) Gordon Ramsay (chef); (c)-(d) The Rolling Stones; (a), (c) object detection, (b), (d) scene recognition.}
  \label{fig:instagram_client_server}
\end{figure*}

Next, we gathered publicly available subsets of first photos ordered by date of several celebrities and companies from Instagram accounts. At first, we compare the results of the client-side models with the proposed pipeline (Fig.~\ref{fig:instagram_client_server}). As one expects, the remote processing of public photos increases the number of detected objects. As a result, the red bars in Fig.~\ref{fig:instagram_client_server}(a),(c) are in most cases larger than the blue bars. However, the whole structure of the gathered profiles is very similar. Thus, if the number of photos in a gallery is not too small, even secure offline processing with lightweight client-side models is appropriate.

\begin{figure*}[t!]
 \centering
  \begin{subfigure}{0.27\textwidth}
    \includegraphics[width=\textwidth, height=0.1\textheight]{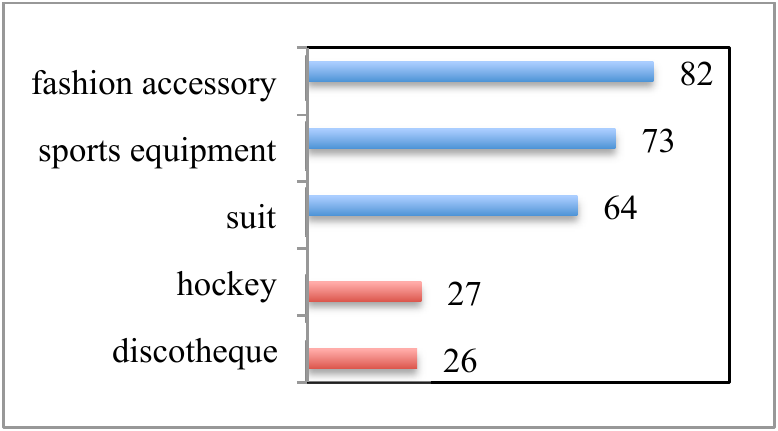}
    \caption{}
  \end{subfigure}
  \begin{subfigure}{0.37\textwidth}
    \includegraphics[width=\textwidth, height=0.1\textheight]{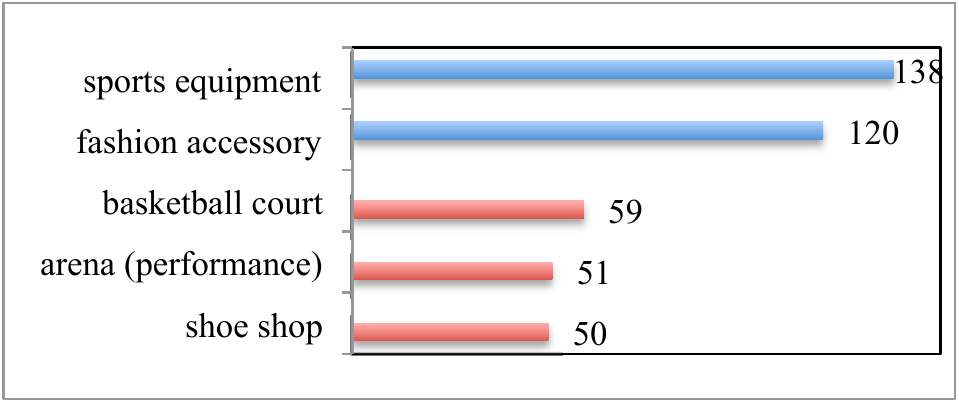}
    \caption{}
  \end{subfigure}
  \begin{subfigure}{0.32\textwidth}
    \includegraphics[height=0.1\textheight]{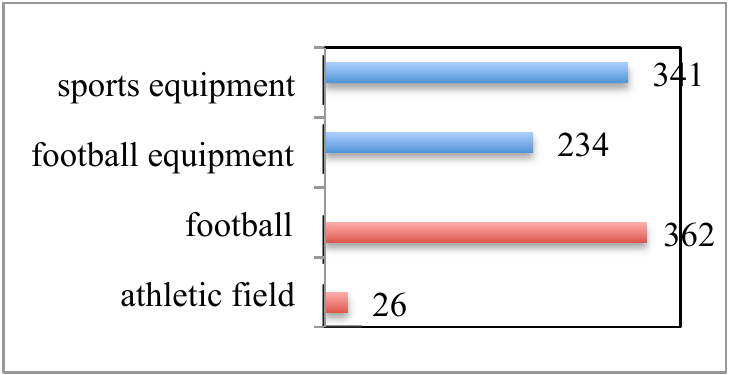}
    \caption{}
  \end{subfigure}
  
  \begin{subfigure}{0.28\textwidth}
    \includegraphics[width=\textwidth, height=0.1\textheight]{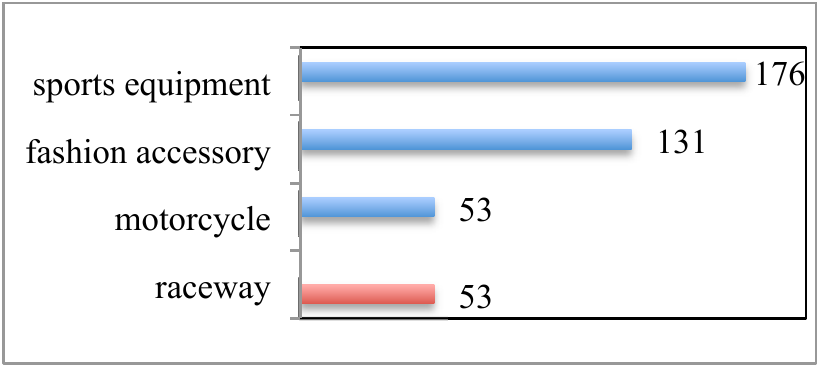}
    \caption{}
  \end{subfigure}
  \begin{subfigure}{0.35\textwidth}
    \includegraphics[height=0.11\textheight]{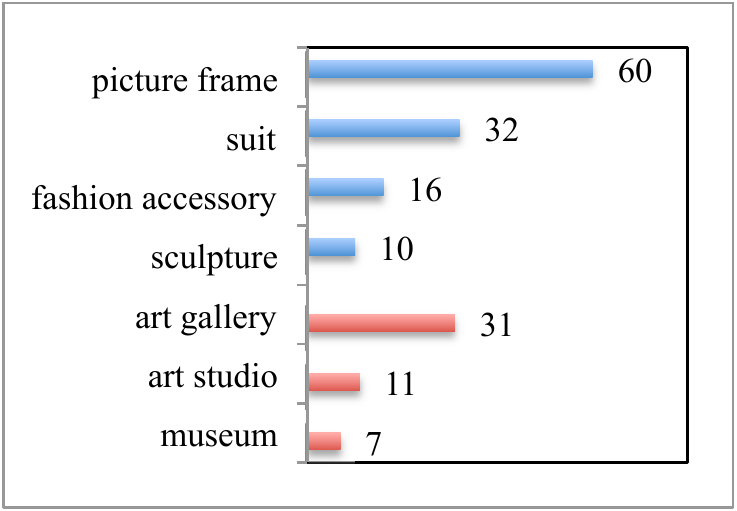}
    \caption{}
  \end{subfigure}
  \begin{subfigure}{0.35\textwidth}
    \includegraphics[height=0.11\textheight]{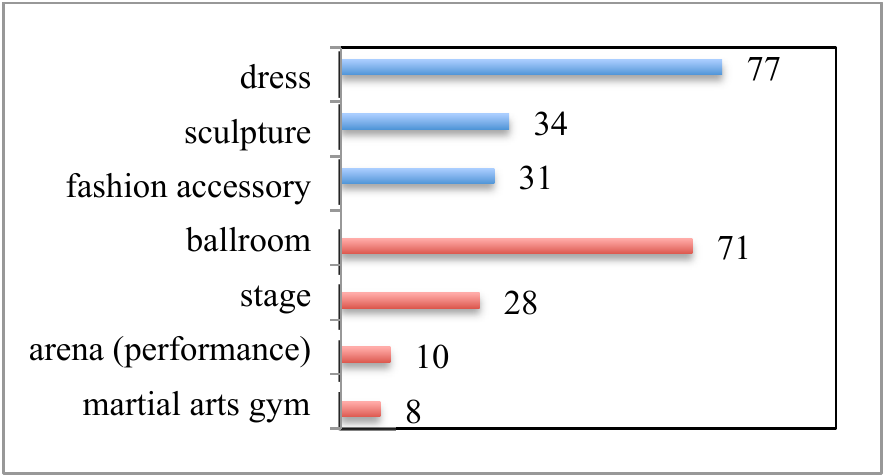}
    \caption{}
  \end{subfigure}
  
  \begin{subfigure}{0.49\textwidth}
    \includegraphics[height=0.17\textheight]{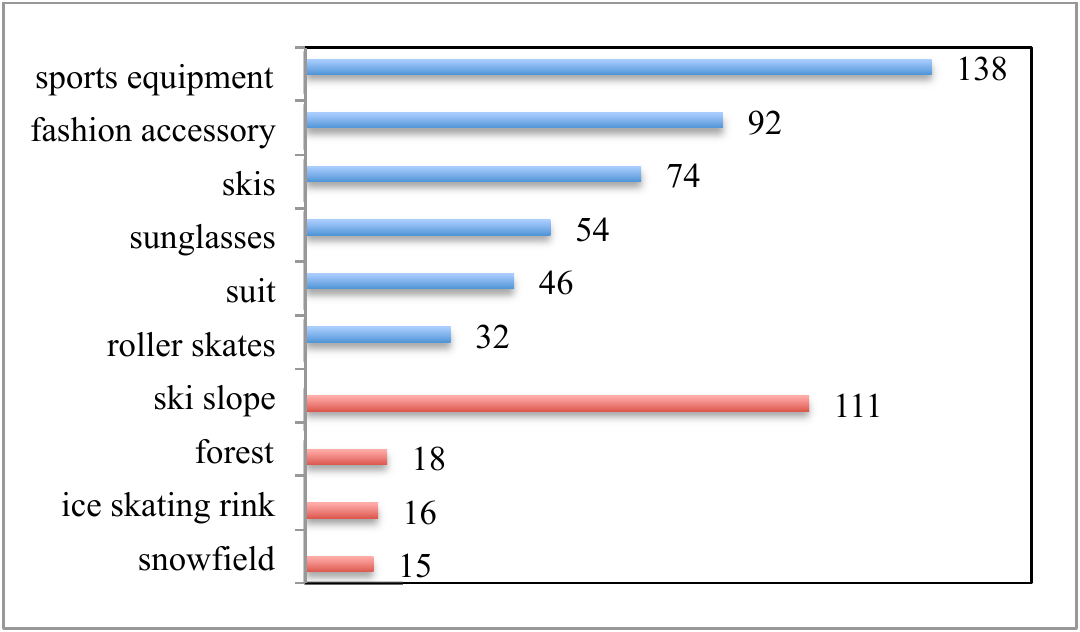}
    \caption{}
  \end{subfigure}
  \begin{subfigure}{0.49\textwidth}
    \includegraphics[height=0.17\textheight]{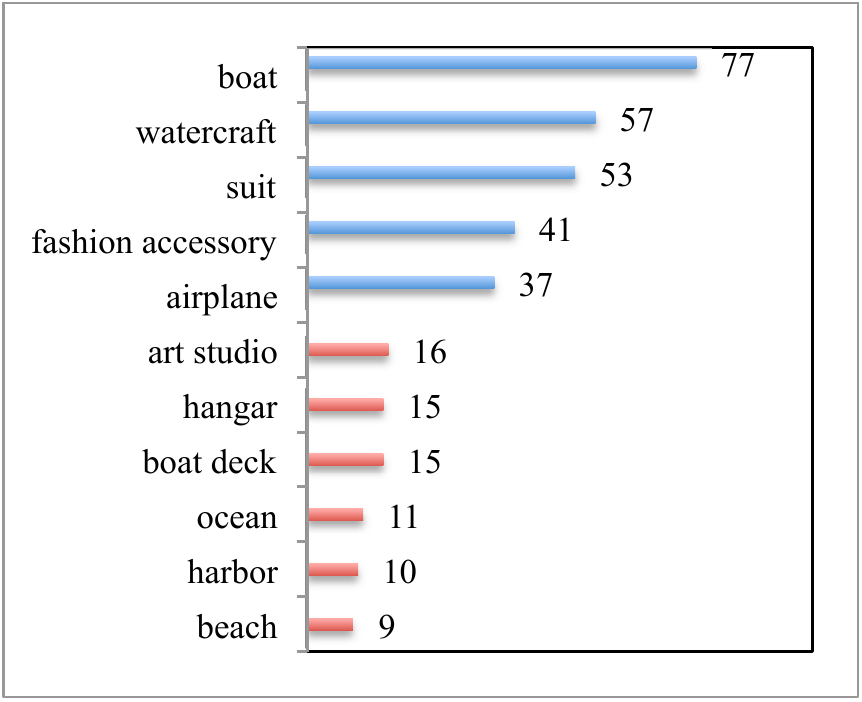}
    \caption{}
  \end{subfigure}
  
  \begin{subfigure}{0.49\textwidth}
    \includegraphics[height=0.17\textheight]{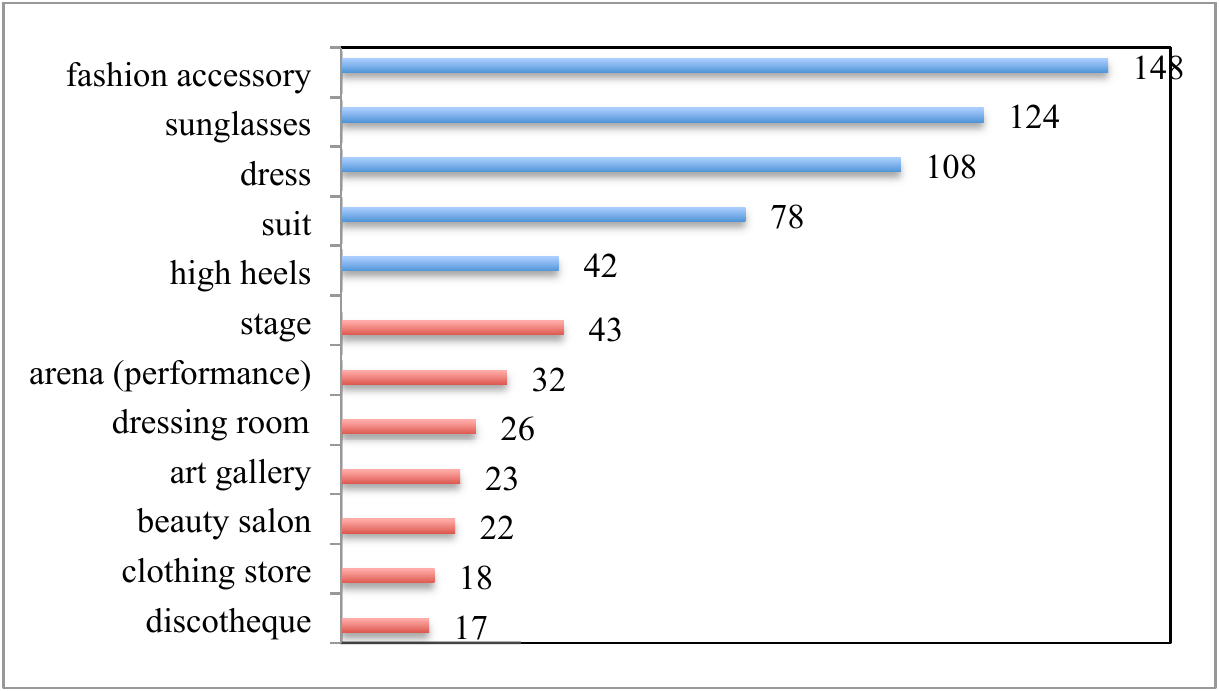}
    \caption{}
  \end{subfigure}
  \begin{subfigure}{0.49\textwidth}
    \includegraphics[ height=0.17\textheight]{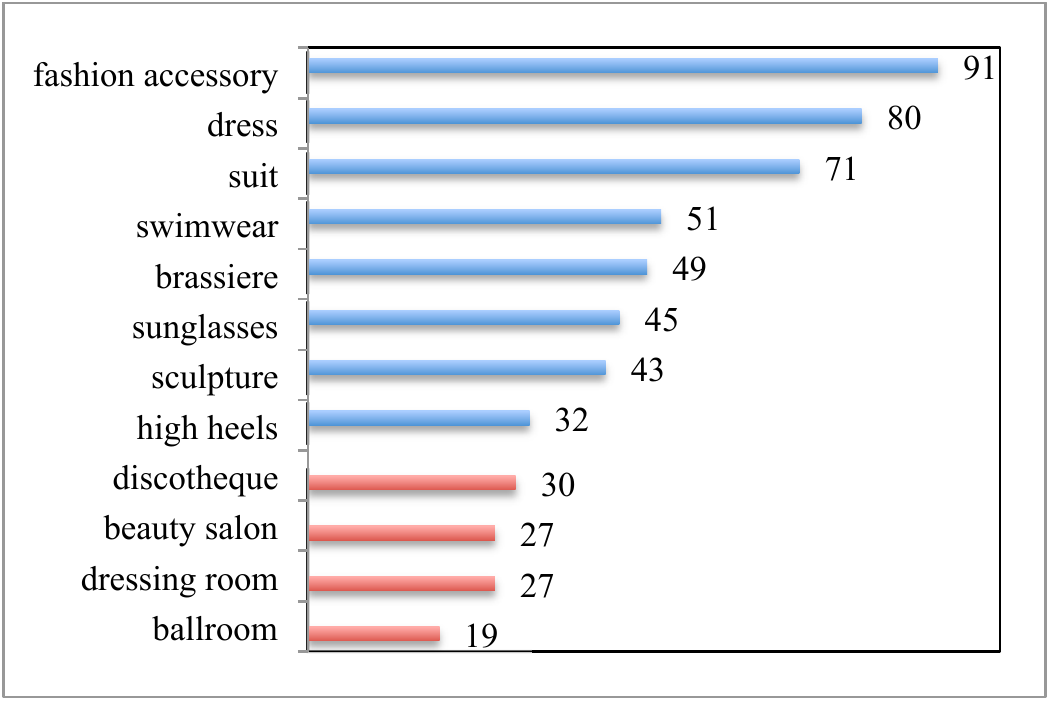}
    \caption{}
  \end{subfigure}
  \caption{Profiles gathered for Instagram photos using client-side models: (a) Alex Ovechkin (hockey player); (b) LeBron James (basketball player); (c) Leo Messi (football player); (d) Max Verstappen (F1 driver); (e) Kehinde Wiley (artist); (g) Svetlana Zakharova (ballet dancer); (h) Johannes Klaebo (skier); (i) Fedor Konyukhov (traveler); (j) Beyonce (singer); (g) Kim Kardashian (media personality).}
  \label{fig:instagram_photos1}
\end{figure*}

\begin{figure*}[t!]
 \centering
  \begin{subfigure}{0.49\textwidth}
    \includegraphics[height=0.16\textheight]{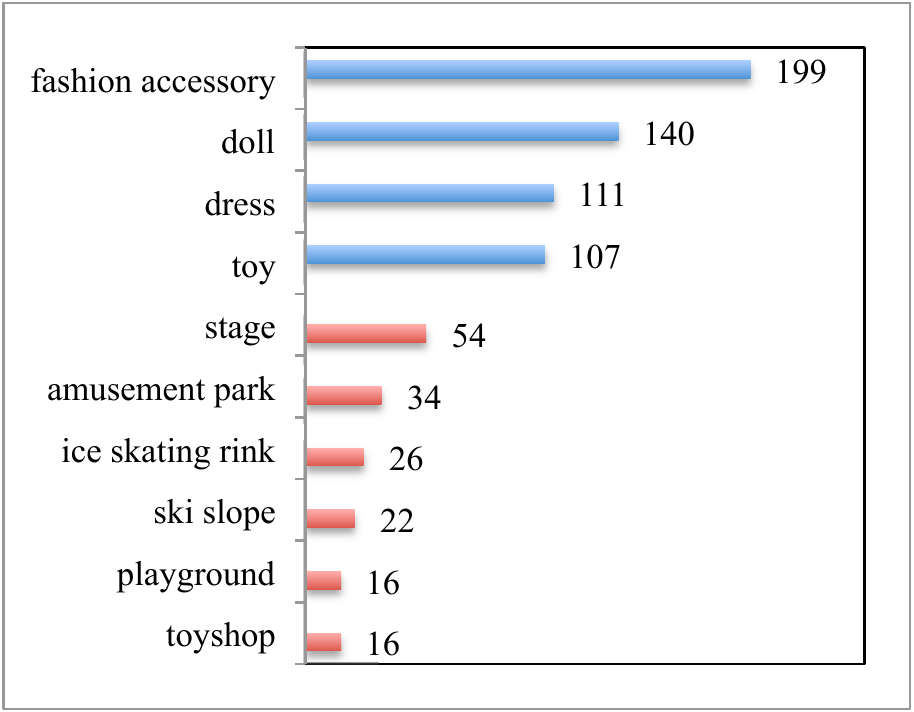}
    \caption{}
  \end{subfigure}
  \begin{subfigure}{0.49\textwidth}
    \includegraphics[height=0.11\textheight]{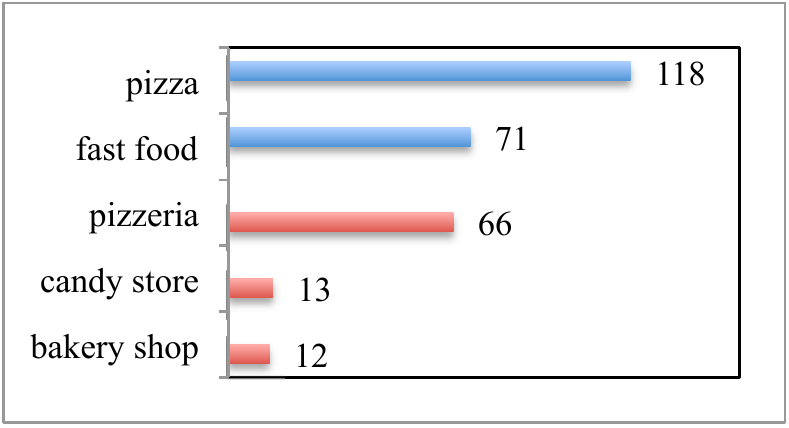}
    \caption{}
  \end{subfigure}
  
  \begin{subfigure}{0.49\textwidth}
    \includegraphics[height=0.13\textheight]{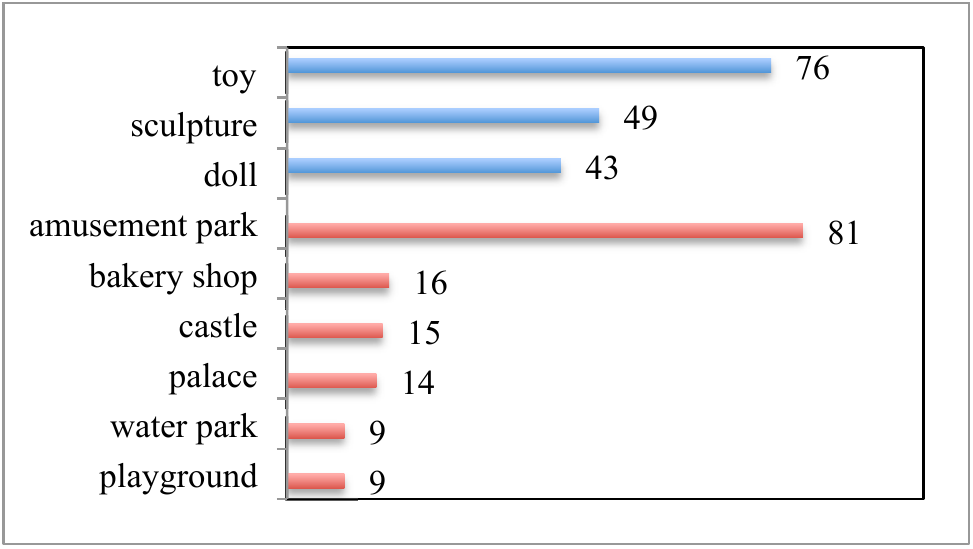}
    \caption{}
  \end{subfigure}
  \begin{subfigure}{0.49\textwidth}
    \includegraphics[height=0.13\textheight]{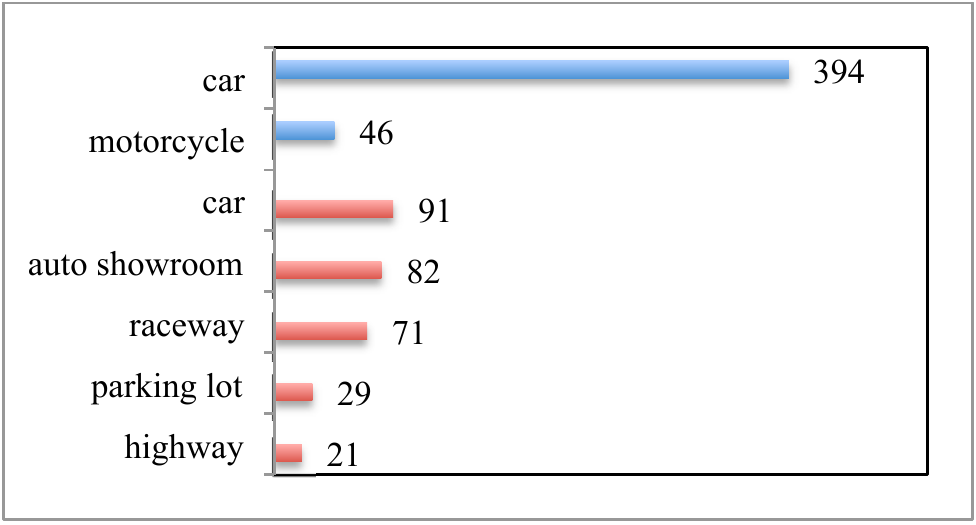}
    \caption{}
  \end{subfigure}

  \begin{subfigure}{0.49\textwidth}
    \includegraphics[height=0.16\textheight]{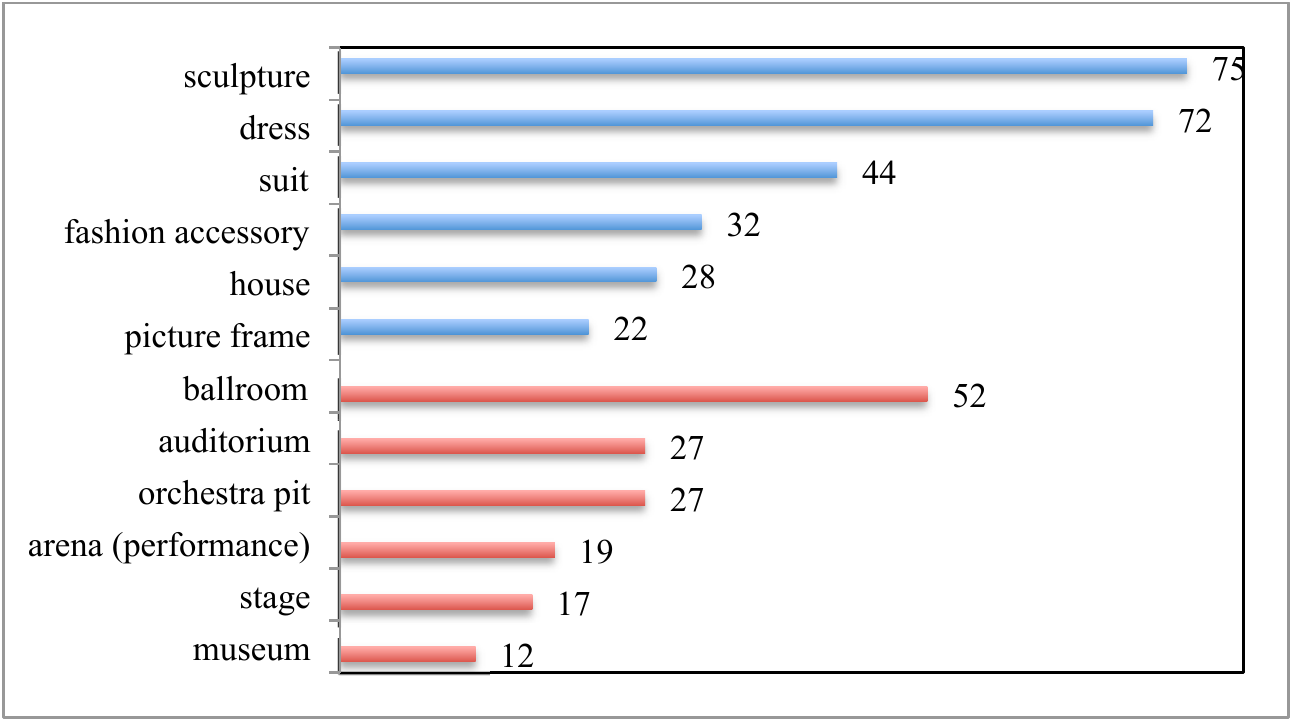}
    \caption{}
  \end{subfigure}
  \begin{subfigure}{0.49\textwidth}
    \includegraphics[width=\textwidth, height=0.16\textheight]{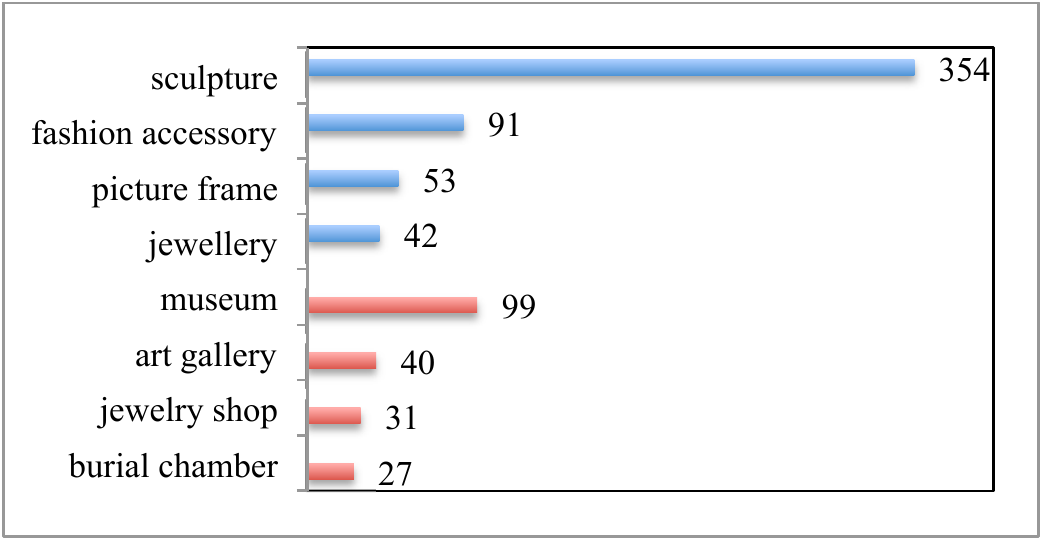}
    \caption{}
  \end{subfigure}
    
  \begin{subfigure}{0.49\textwidth}
    \includegraphics[width=\textwidth, height=0.2\textheight]{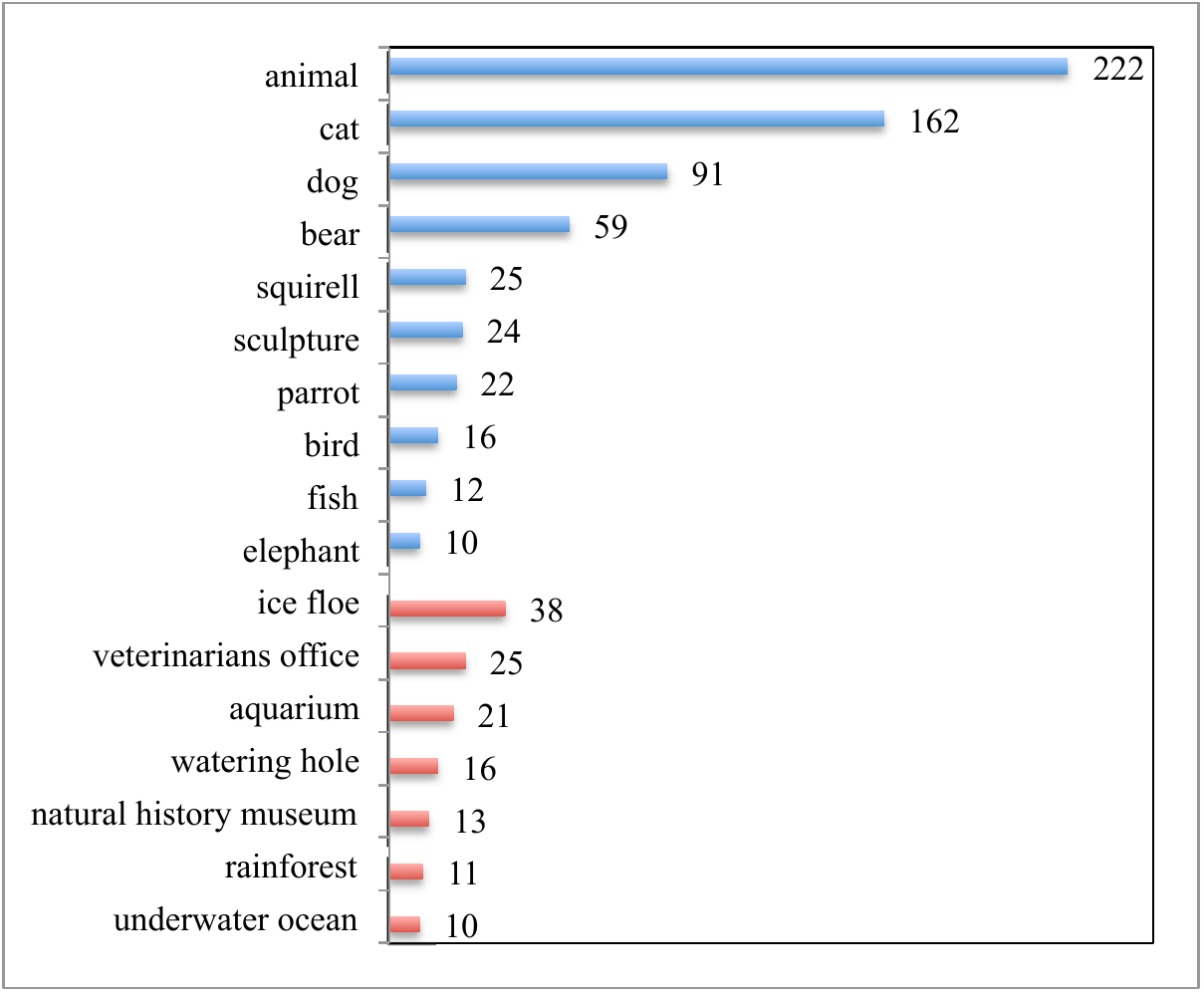}
    \caption{}
  \end{subfigure}
  \begin{subfigure}{0.49\textwidth}
    \includegraphics[width=\textwidth, height=0.2\textheight]{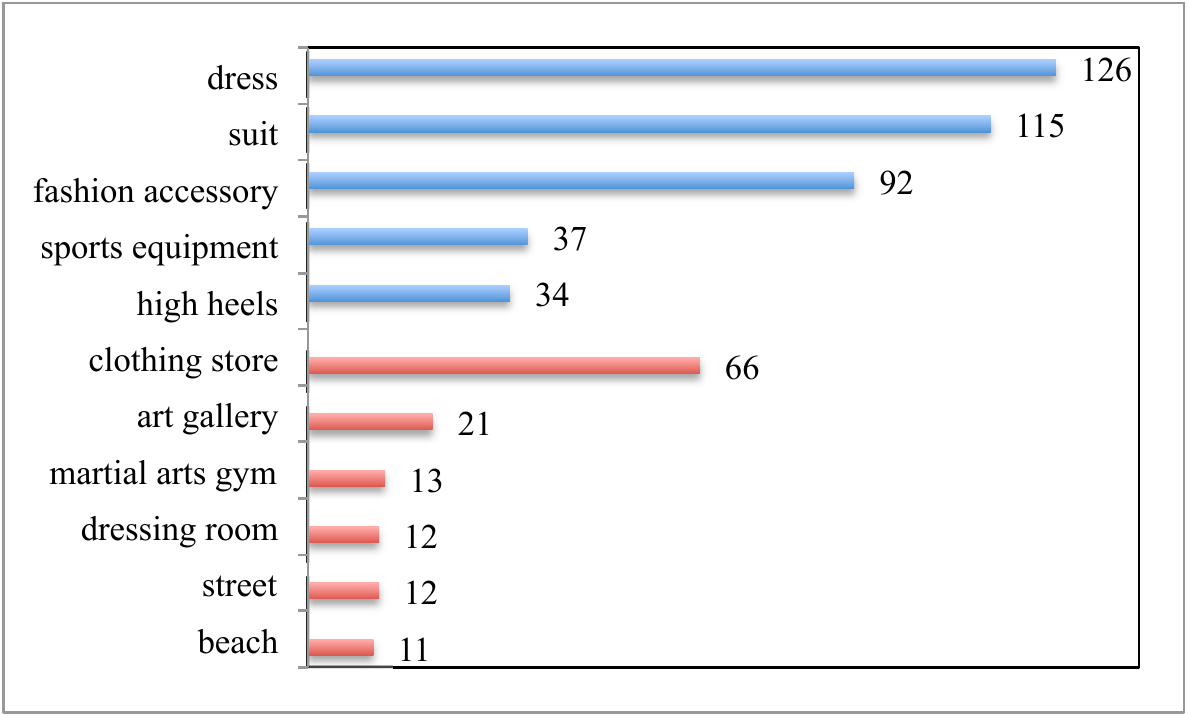}
    \caption{}
  \end{subfigure}
  \caption{Profiles gathered for Instagram photos using client-side models: (a) Ded Moroz (Santa); (b) Mir Pizzy (pizzeria); (c) Disneyland; (d) Ferrari; (e) Bolshoi theater; (f) Louvre; (g) Moscow zoo; (h) Uniqlo.}
  \label{fig:instagram_photos2}
\end{figure*}

Finally, other results of client-side models are shown in Fig.~\ref{fig:instagram_photos1} and Fig.~\ref{fig:instagram_photos2}. They are even more attractive when compared to the first examples (Fig.~\ref{fig:web_photos}). Again, scene recognition characterizes the subject almost perfectly, and the detected objects contain important information for scene and event processing. It is necessary to highlight that the largest facial cluster obtained in our pipeline for all cases of single person account contains the photo of this person. The second largest clusters contain the main relatives (wife/boyfriend, children, etc.). The face clustering~\cite{savchenko2019efficient} of the Rolling stones gallery resulted to 4 large clusters of the main musicians of the rock band.

\subsection{Discussion} \label{subsec:4.5}

The usage of the proposed pipeline (Fig.~\ref{fig:final_pipeline}) in various image recognition tasks leads to the following main results:

\begin{enumerate}
\item Developed structurally pruned MobileNet-based scene recognition model has near state-of-the-art accuracy for classification of more than 300 different scenes and requires only 60-85 ms to process an image on mobile device (Table~\ref{table:scenerec_places_results}).
\item We improved the previous state-of-the-art for PEC from 62.2\%~\cite{wang2018transferring} even for client-side model (63.34\%). Our server-side model is even better (64.98\%).
\item The accuracy of our approach for the WIDER dataset is 7.4-9.3\% higher when compared to the best results (42.4\%) from original paper~\cite{xiong2015recognize}.
\item The proposed aggregation of visual features in an album improves the state-of-the-art results for the original testing protocol for the PEC dataset~\cite{bossard2013event} by using the client-side models (Table~\ref{table:pec_attention}).
\end{enumerate}

Our approach overcomes some key privacy issues that are implicit in computations performed on remote servers. Despite typical applications of private (sensitive) image detection~\cite{tran2016privacy}, our server-side processing does not need to store the images in any database as it immediately removes any input photo right after processing in ``Accurate scene classification" and ``Accurate object detection" modules. Thus, it is vulnerable only to network sniffing attacks. In practical applications it is important to add traffic encryption and/or implement a server as a part of cloud storages (Samsung Cloud, Apple iCloud, etc.). However, we should emphasize that any above-mentioned heuristic (detection of faces and sensitive words) may fail in some specific cases because it is impossible to \textit{automatically} detect \textit{all} private images with 100\% true positive rate and non-zero true negative rate. Hence, in order to improve the generalizability of the proposed method, we provide several opportunities for a user. If the above-mentioned guarantees with the prohibition of storage of all photos in an input server and our privacy detection engine are appropriate, the whole pipeline (Fig.~\ref{fig:final_pipeline}) is used. However, if the privacy is extremely important for a particular user, we let him to explicitly prevent privacy detection. In such case all photos are marked as private in order to process the whole gallery in a mobile device. This option is chosen by default. If the user verifies that our engine predicts private/public status correctly, he could turn on the option of remote processing of public photos.

Certainly, though the client-side models let us obtain very accurate profile for many examples of Instagram accounts (Fig.~\ref{fig:instagram_client_server}, Fig.~\ref{fig:instagram_photos1} and Fig.~\ref{fig:instagram_photos2}), predicted user's profile will be less accurate when compared to the whole pipeline. However, our experimental study demonstrated that the scene recognition results of MobileNet scores and features for both event datasets from Subsection\ref{subsec:4.2} are only 0.2-2.15\% less accurate when compared to Inception model. In this case there is no practical need for scene classification on remote server. However, the situation is different for object detection task due to the limitations of SSD-based models to detect small objects (food, pets, fashion accessories, etc.). Indeed, though the processing in client-mode is better than the random guess with accuracy $100\%/14 \approx 7.14\%$ for PEC and $100\%/61 \approx 1.64\%$ for WIDER, the accuracy of SSDLite are much (6-13\%) lower than the accuracy of Faster R-CNN. In this particular example the proposed representations hides this disadvantage and the best ensemble for the client-side models is not too worth than the server-side models (63.34\% vs 64.98\% for PEC and 49.80\% vs 51.76\% for WIDER). However, in general, the user's profile gathered with Faster R-CNN detectors has much more objects so that the remote processing of public photos is worth implementing.

\section{Conclusion}
\label{sec:5}

In this paper, we have proposed novel pipeline (Fig.~\ref{fig:final_pipeline}) for recognition of user's preferences for hobbies and lifestyle in visual data based on the representation of the photos with scores and/or embeddings of scene recognition model and outputs of object detectors. It was demonstrated how to efficiently combine the same features of each photo in a given input set into a single descriptor of particular user (\ref{eq:squeeze_operator})-(\ref{eq:squeezed_weights}) based on the learnable pooling used previously only for video recognition~\cite{yang2017neural,miech2017learnable}. We achieved the state-of-the-art results for Photo Event Collection~\cite{bossard2013event} even by using the client-side models (Tables~\ref{table:wider_mobilenet},~\ref{table:pec_attention}). 

Our engine was implemented in the publicly available Android application (Fig.~\ref{fig:mobile_demo_user_profile},~\ref{fig:mobile_demo_details}). It is applicable for various personalized mobile services such as target advertisements, marketing and recommender systems. For example, users can get better personalization while traveling. Depending on the user's profile, it is possible to recommend suitable products, shops, content. The cold-start problem could be alleviated if the companies can get access to reliable preference information without the need of rating elicitation of monitoring users closely over time. 

One of the main limitations of the proposed Algorithm~\ref{algorithm2} is the need to perform rather slow object detection. If only scenes/events are needed in a user's profile, preliminarily object detection can significantly increase the processing time. Moreover, we have not still reached the state-of-the-art accuracy (53\%) for the WIDER dataset~\cite{wang2018transferring}. Our best approach is 1.24\% less accurate (Table~\ref{table:wider_mobilenet}). Finally, there is a possibility of having some images to be misclassified as public and sent to the remote server, which may be undesirable for very secure applications even if it is guaranteed that the images will be removed there right after scene classification and object detection and the channel is encrypted. Though a user can turn off the remote processing completely, it may reduce the total number of detected objects in his or her profile (Fig.~\ref{fig:instagram_client_server}).

In future it is important to improve the accuracy of private photo detection~\cite{tran2016privacy} by gathering large datasets and training classifiers with manageable false positive rate. Secondly, it is necessary to apply the transfer learning techniques from~\cite{wang2018transferring} to our representations in order to increase the accuracy for the WIDER dataset. Finally, it is desirable to extend the number of categories by estimating concrete characteristics of extracted objects, e.g., pet breeds, car models, sport teams, logos~\cite{su2020scalable}, etc. 

\section*{Acknowledgement}
This research is based on the work supported by Samsung Research, Samsung Electronics. The work of A.V. Savchenko is supported by the National Research University Higher School of Economics (HSE) within the framework of Basic Research Program.
\section*{References}

\bibliography{savchenko}

\begin{thebibliography}{10}
\expandafter\ifx\csname url\endcsname\relax
  \def\url#1{\texttt{#1}}\fi
\expandafter\ifx\csname urlprefix\endcsname\relax\def\urlprefix{URL }\fi
\expandafter\ifx\csname href\endcsname\relax
  \def\href#1#2{#2} \def\path#1{#1}\fi

\bibitem{yu2019cross}
X.~Yu, F.~Jiang, J.~Du, D.~Gong, A cross-domain collaborative filtering
  algorithm with expanding user and item features via the latent factor space
  of auxiliary domains, Pattern Recognition 94 (2019) 96--109.

\bibitem{yang2020graph}
G.~Yang, J.~Cao, Z.~Chen, J.~Guo, J.~Li, Graph-based neural networks for
  explainable image privacy inference, Pattern Recognition 105 (2020) 107360.

\bibitem{goodfellow2016deep}
I.~Goodfellow, Y.~Bengio, A.~Courville, Deep learning, MIT press, 2016.

\bibitem{farinella2015representing}
G.~M. Farinella, D.~Rav{\`\i}, V.~Tomaselli, M.~Guarnera, S.~Battiato,
  Representing scenes for real-time context classification on mobile devices,
  Pattern Recognition 48~(4) (2015) 1086--1100.

\bibitem{zhou2018places}
B.~Zhou, A.~Lapedriza, A.~Khosla, A.~Oliva, A.~Torralba, Places: A 10 million
  image database for scene recognition, IEEE Transactions on Pattern Analysis
  and Machine Intelligence 40~(6) (2018) 1452--1464.

\bibitem{huang2017speed}
J.~Huang, V.~Rathod, C.~Sun, M.~Zhu, A.~Korattikara, A.~Fathi, I.~Fischer,
  Z.~Wojna, Y.~Song, S.~Guadarrama, Speed/accuracy trade-offs for modern
  convolutional object detectors, in: Proceedings of the Conference on Computer
  Vision and Pattern Recognition (CVPR), IEEE, 2017, pp. 7310--7311.

\bibitem{grechikhin2019user}
I.~Grechikhin, A.~V. Savchenko, User modeling on mobile device based on facial
  clustering and object detection in photos and videos, in: Iberian Conference
  on Pattern Recognition and Image Analysis (IbPRIA), Springer, 2019, pp.
  429--440.

\bibitem{su2020scalable}
H.~Su, S.~Gong, X.~Zhu, Scalable logo detection by self co-learning, Pattern
  Recognition 97 (2020) 107003.

\bibitem{savchenko2019efficient}
A.~V. Savchenko, Efficient facial representations for age, gender and identity
  recognition in organizing photo albums using multi-output {ConvNet}, PeerJ
  Computer Science 5~(e197).
\newblock \href {http://dx.doi.org/10.7717/peerj-cs.197}
  {\path{doi:10.7717/peerj-cs.197}}.

\bibitem{wang2018transferring}
L.~Wang, Z.~Wang, Y.~Qiao, L.~Van~Gool, Transferring deep object and scene
  representations for event recognition in still images, International Journal
  of Computer Vision 126~(2-4) (2018) 390--409.

\bibitem{yang2017neural}
J.~Yang, P.~Ren, D.~Zhang, D.~Chen, F.~Wen, H.~Li, G.~Hua, Neural aggregation
  network for video face recognition, in: Proceedings of the International
  Conference on Computer Vision and Pattern Recognition (CVPR), IEEE, 2017, pp.
  5216--5225.

\bibitem{szegedy2017inception}
C.~Szegedy, S.~Ioffe, V.~Vanhoucke, A.~A. Alemi, Inception-v4,
  {Inception-ResNet} and the impact of residual connections on learning, in:
  Proceedings of the AAAI Conference on Artificial Intelligence, 2017, pp.
  4278--4284.

\bibitem{han2015deep}
S.~Han, H.~Mao, W.~J. Dally, Deep compression: Compressing deep neural networks
  with pruning, trained quantization and {Huffman} coding, in: Proceedings of
  International Conference on Learning Representations (ICLR), 2016.

\bibitem{mittal2018recovering}
D.~Mittal, S.~Bhardwaj, M.~M. Khapra, B.~Ravindran, Recovering from random
  pruning: On the plasticity of deep convolutional neural networks, in:
  Proceedings of Winter Conference on Applications of Computer Vision (WACV),
  IEEE, 2018, pp. 848--857.

\bibitem{molchanov2016pruning}
P.~Molchanov, S.~Tyree, T.~Karras, T.~Aila, J.~Kautz, Pruning convolutional
  neural networks for resource efficient inference, arXiv preprint
  arXiv:1611.06440.

\bibitem{grachev2019}
A.~M. Grachev, D.~I. Ignatov, A.~V. Savchenko, Compression of recurrent neural
  networks for efficient language modeling, Applied Soft Computing 79 (2019)
  354--362.

\bibitem{rothe2015dldr}
R.~Rothe, R.~Timofte, L.~Van~Gool, Dldr: Deep linear discriminative retrieval
  for cultural event classification from a single image, in: Proceedings of the
  International Conference on Computer Vision Workshops (ICCVW), 2015, pp.
  53--60.

\bibitem{ssd}
W.~Liu, D.~Anguelov, D.~Erhan, C.~Szegedy, S.~Reed, C.~Y. Fu, A.~C. Berg,
  {SSD}: Single shot multibox detector, in: Proceedings of European Conference
  on Computer Vision (ECCV), Springer, Cham, 2016, pp. 21--37.

\bibitem{sandler_inverted_2018}
M.~Sandler, A.~Howard, M.~Zhu, A.~Zhmoginov, L.-C. Chen, {MobilenetV2}:
  Inverted residuals and linear bottlenecks, in: Proceedings of the Conference
  on Computer Vision and Pattern Recognition (CVPR), 2018, pp. 4510--4520.

\bibitem{ren2015faster}
S.~Ren, K.~He, R.~Girshick, J.~Sun, Faster {R-CNN}: Towards real-time object
  detection with region proposal networks, in: Advances in Neural Information
  Processing Systems (NIPS), 2015, pp. 91--99.

\bibitem{xiong2015recognize}
Y.~Xiong, K.~Zhu, D.~Lin, X.~Tang, Recognize complex events from static images
  by fusing deep channels, in: Proceedings of the International Conference on
  Computer Vision and Pattern Recognition (CVPR), 2015, pp. 1600--1609.

\bibitem{you2016picture}
Q.~You, S.~Bhatia, J.~Luo, A picture tells a thousand words about you! {User}
  interest profiling from user generated visual content, Signal Processing 124
  (2016) 45--53.

\bibitem{chen2016context}
T.~Chen, X.~He, M.-Y. Kan, Context-aware image tweet modelling and
  recommendation, in: Proceedings of the 24th International Conference on
  Multimedia, ACM, 2016, pp. 1018--1027.

\bibitem{deldjoo2016content}
Y.~Deldjoo, M.~Elahi, P.~Cremonesi, F.~Garzotto, P.~Piazzolla, M.~Quadrana,
  Content-based video recommendation system based on stylistic visual features,
  Journal on Data Semantics 5~(2) (2016) 99--113.

\bibitem{shankar2017deep}
D.~Shankar, S.~Narumanchi, H.~Ananya, P.~Kompalli, K.~Chaudhury, Deep learning
  based large scale visual recommendation and search for e-commerce, arXiv
  preprint arXiv:1703.02344.

\bibitem{andreeva2018extraction}
E.~Andreeva, D.~I. Ignatov, A.~Grachev, A.~V. Savchenko, Extraction of visual
  features for recommendation of products via deep learning, in: Proceedings of
  International Conference on Analysis of Images, Social Networks and Texts
  (AIST), Springer, 2018, pp. 201--210.

\bibitem{zhai2017visual}
A.~Zhai, D.~Kislyuk, Y.~Jing, M.~Feng, E.~Tzeng, J.~Donahue, Y.~L. Du,
  T.~Darrell, Visual discovery at {PInterest}, in: Proceedings of the 26th
  International Conference on World Wide Web Companion, 2017, pp. 515--524.

\bibitem{kopeykina2019automatic}
L.~N. Kopeykina, A.~V. Savchenko, Automatic privacy detection in scanned
  document images based on deep neural networks, in: Proceedings of
  International Russian Automation Conference (RusAutoCon), IEEE, 2019.

\bibitem{tran2016privacy}
L.~Tran, D.~Kong, H.~Jin, J.~Liu, Privacy-{CNH}: A framework to detect photo
  privacy with convolutional neural network using hierarchical features, in:
  Proceedings of the AAAI Conference on Artificial Intelligence, 2016.

\bibitem{savchenko2017maximum}
A.~V. Savchenko, Maximum-likelihood approximate nearest neighbor method in
  real-time image recognition, Pattern Recognition 61 (2017) 459--469.

\bibitem{savchenko2018unconstrained}
A.~V. Savchenko, N.~S. Belova, Unconstrained face identification using maximum
  likelihood of distances between deep off-the-shelf features, Expert Systems
  with Applications 108 (2018) 170--182.

\bibitem{cao2018vggface2}
Q.~Cao, L.~Shen, W.~Xie, O.~M. Parkhi, A.~Zisserman, Vggface2: A dataset for
  recognising faces across pose and age, in: Proceedings of the 13th IEEE
  International Conference on Automatic Face \& Gesture Recognition (FG), IEEE,
  2018, pp. 67--74.

\bibitem{savchenko2018efficient}
A.~V. Savchenko, Efficient statistical face recognition using trigonometric
  series and cnn features, in: Proceedings of the 24th International Conference
  on Pattern Recognition (ICPR), IEEE, 2018, pp. 3262--3267.

\bibitem{rassadin2019scene}
A.~G. Rassadin, A.~V. Savchenko, Scene recognition in user preference
  prediction based on classification of deep embeddings and object detection,
  in: Proceedings of the International Symposium on Neural Networks (ISNN),
  Springer, 2019, pp. 422--430.

\bibitem{bossard2013event}
L.~Bossard, M.~Guillaumin, L.~Van~Gool, Event recognition in photo collections
  with a stopwatch {HMM}, in: Proceedings of the International Conference on
  Computer Vision (ICCV), IEEE, 2013, pp. 1193--1200.

\bibitem{kang2017visually}
W.-C. Kang, C.~Fang, Z.~Wang, J.~McAuley, Visually-aware fashion recommendation
  and design with generative image models, in: Proceedings of International
  Conference on Data Mining (ICDM), IEEE, 2017, pp. 207--216.

\bibitem{miech2017learnable}
A.~Miech, I.~Laptev, J.~Sivic, Learnable pooling with context gating for video
  classification, arXiv preprint arXiv:1706.06905.

\bibitem{wang2017recognizing}
Y.~Wang, Z.~Lin, X.~Shen, R.~Mech, G.~Miller, G.~W. Cottrell, Recognizing and
  curating photo albums via event-specific image importance, in: Proceedings of
  British Conference on Machine Vision (BMVC), 2017.

\bibitem{wu2015learning}
Z.~Wu, Y.~Huang, L.~Wang, Learning representative deep features for image set
  analysis, IEEE Transactions on Multimedia 17~(11) (2015) 1960--1968.

\end{thebibliography}

\end{document}